\documentclass[10pt,twocolumn,letterpaper]{article}

\usepackage[pagenumbers]{cvpr} 

\usepackage{graphicx}
\usepackage{amsmath}
\usepackage{amssymb}
\usepackage{booktabs}
\usepackage[accsupp]{axessibility}

\usepackage{balance}

\usepackage{amsmath,amsfonts,bm}

\def\eqref#1{equation~\ref{#1}}

\def\1{\bm{1}}

\def\vw{{\bm{w}}}
\def\vx{{\bm{x}}}
\def\vy{{\bm{y}}}
\def\vz{{\bm{z}}}

\def\mB{{\bm{B}}}

\def\mI{{\bm{I}}}

\def\mV{{\bm{V}}}
\def\mW{{\bm{W}}}

\DeclareMathAlphabet{\mathsfit}{\encodingdefault}{\sfdefault}{m}{sl}
\SetMathAlphabet{\mathsfit}{bold}{\encodingdefault}{\sfdefault}{bx}{n}

\def\gG{{\mathcal{G}}}

\def\gL{{\mathcal{L}}}

\def\gN{{\mathcal{N}}}

\usepackage{amsthm}
\usepackage{multirow}       
\usepackage{wrapfig}        
\usepackage{xcolor}
\usepackage{enumitem}
\usepackage{threeparttable}

\usepackage{algorithm}   
\usepackage[noend]{algpseudocode}

\usepackage[utf8]{inputenc}
\usepackage{kotex}

\newcommand{\genie}{\textsc{Genie}}
\newcommand{\citep}{\cite}
\usepackage{mathtools}
\usepackage{multirow}
\usepackage{pifont}
\usepackage{tabularx}
\usepackage{amssymb}
\usepackage{booktabs}
\usepackage{color}
\usepackage{commath}
\usepackage{float}
\usepackage{rotating}
\usepackage{arydshln}
\usepackage{soul}
\newcommand{\revise}{\textcolor{black}}

\newcommand{\bns}{\mathcal{L}^\mathrm{D}_\mathrm{BNS}}

\usepackage{hyperref}
\hypersetup{colorlinks,breaklinks}
\usepackage{etoolbox}
\makeatletter
\let\@titlehook=\relax
\apptocmd{\@maketitle}{\@titlehook}{}{}
\newcommand{\titlehook}[1]{\def\@titlehook{#1}}
\makeatother

\usepackage[capitalize]{cleveref}
\crefname{section}{Sec.}{Secs.}
\Crefname{section}{Section}{Sections}
\Crefname{table}{Table}{Tables}
\crefname{table}{Tab.}{Tabs.}
\crefname{figure}{Figure}{Figures}

    \makeatletter
\def\@fnsymbol#1{\ensuremath{\ifcase#1\or *\or *\or
   *\or \mathparagraph\or \|\or **\or \dagger\dagger
   \or \ddagger\ddagger \else\@ctrerr\fi}}
    \makeatother

\def\cvprPaperID{1783} 
\def\confName{CVPR}
\def\confYear{2023}

\begin{document}

\title{\textsc{Genie}: Show Me the Data for Quantization}
\author{Yongkweon Jeon\thanks{Equal contribution. Correspondence to: \textit{dragwon.jeon@samsung.com}}~~~~~~~~Chungman Lee$^*$~~~~~~~~Ho-young Kim$^*$\\
\normalsize Samsung Research\\
{\tt\small \{dragwon.jeon, chungman.lee, hoyoung4.kim\}@samsung.com}
}

\titlehook{
    \begin{center}
        \captionsetup{type=figure}
        \includegraphics[width=0.92\linewidth]{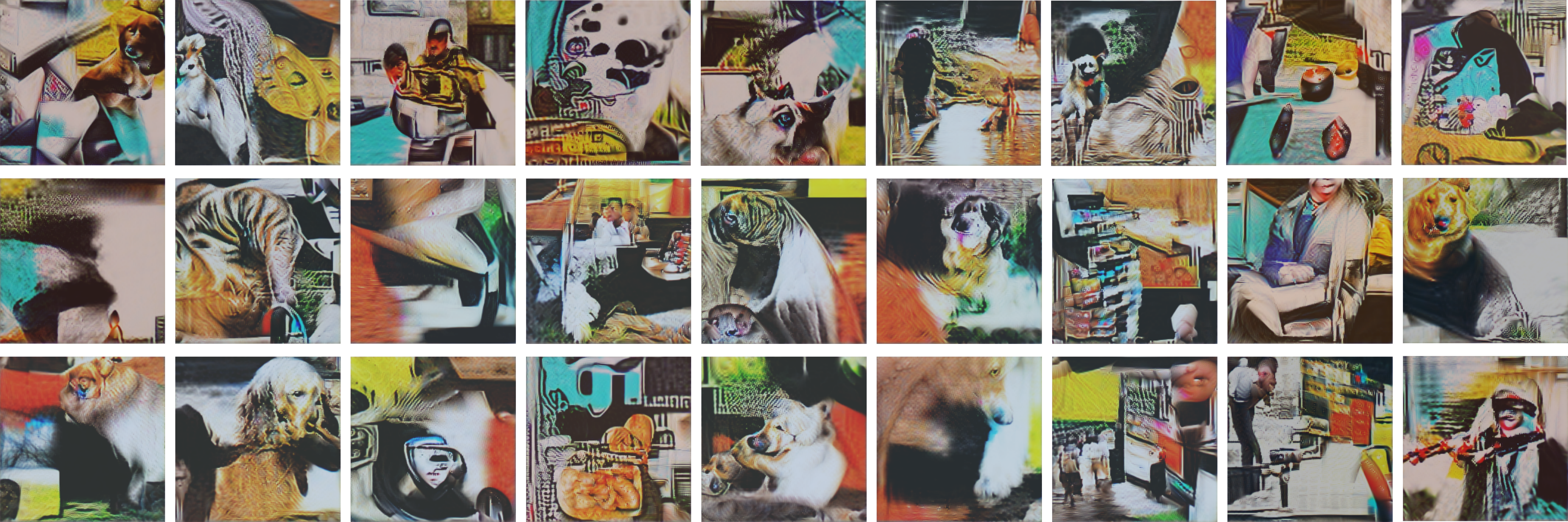}
        \captionof{figure}{Distilled images by \genie~(without any image prior loss).}\label{fig:fig1}
    \end{center}
    }
\maketitle
{\def\iand{\\[5pt]\let\and=\nand}%
 \def\nand{\ifhmode\unskip\nobreak\fi\ $\cdot$ }%
 \let\and=\nand
 \def\at{\\\let\and=\iand}%
\vspace{5dd}}

\begin{abstract}
    Zero-shot quantization is a promising approach for developing lightweight deep neural networks when data is inaccessible owing to various reasons, including cost and issues related to privacy. By exploiting the learned parameters ($\mu$ and $\sigma$) of batch normalization layers in an FP32-pre-trained model, zero-shot quantization schemes focus on generating synthetic data. 
    Subsequently, they distill knowledge from the pre-trained model (\textit{teacher}) to the quantized model (\textit{student}) such that the quantized model can be optimized with the synthetic dataset. 
    However, thus far, zero-shot quantization has primarily been discussed in the context of quantization-aware training methods, which require task-specific losses and long-term optimization as much as retraining. 
    We thus introduce a post-training quantization scheme for zero-shot quantization that produces high-quality quantized networks within a few hours.
    Furthermore, we propose a framework called \genie~that generates data suited for quantization. With the data synthesized by \genie, we can produce robust quantized models without real datasets, which is comparable to few-shot quantization. We also propose a post-training quantization algorithm to enhance the performance of quantized models. By combining them, we can bridge the gap between zero-shot and few-shot quantization while significantly improving the quantization performance compared to that of existing approaches. In other words, we can obtain a unique state-of-the-art zero-shot quantization approach. 
    The code is available at \url{https://github.com/SamsungLabs/Genie}.
\end{abstract}
\section{Introduction}
\label{sec:intro}
Quantization is an indispensable procedure for deploying models in resource-constrained devices such as mobile phones. By representing tensors using a lower bit width and maintaining a dense format of tensors, quantization reduces a computing unit to a significantly smaller size compared to that achieved by other approaches (such as pruning and low-rank approximations) and facilitates massive data parallelism with vector processing units. 
Most early studies utilized quantization-aware training (QAT) schemes~\citep{esser2019learned,nagel22overcoming} to compress models, which requires the entire training dataset and takes as much time as training FP32 models. 
However, access to the entire dataset for quantizing models may not be possible in the real world or industry owing to a variety of reasons, including issues related to privacy preservation. 
Thus, recent studies have emphasized post-training quantization (PTQ)~\citep{nagel2020up,li2021brecq,hubara2021accurate,jeon2022mr} because it serves as a convenient method of producing high-quality quantized networks with only a small amount of unlabeled datasets or even in the absence of a dataset (including synthetic datasets). Because PTQ can compress models within a few hours but shows comparable performance to QAT, PTQ is preferred over QAT in practical situations.

\begin{figure}[t]
    \centering
    \includegraphics[width=.70\linewidth]{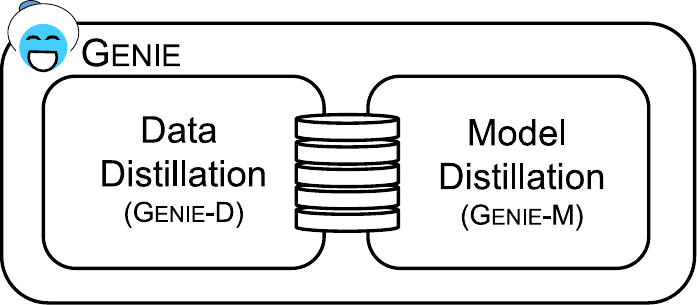}
    \caption{Conceptual illustration of \genie, which consists of two sub-modules: synthesizing data and quantizing models}
    \label{fig:genie}
\end{figure}

Zero-shot quantization (ZSQ)~\citep{liu2021zero,cai2020zeroq,choi2022s} is another research regime that synthesizes data to compress models without employing real datasets. Starting from DFQ~\citep{nagel2019data}, schemes for ZSQ gradually pay more attention to generating elaborate replicas such that the distribution of intermediate feature maps matches the statistics of the corresponding batch normalization layers.
Although many studies have achieved significant advancement in regards to quantization in the absence of real data, most of them have relied on QAT schemes that require task-specific loss, such as \textit{cross-entropy (CE) loss} and \textit{Kullback--Leibler (KL) divergence} ~\citep{kullback1951information}, which requires more than 10 hours to complete the quantization of ResNet-18~\cite{he2016deep} on Nvidia V100.

Excluding the data used, ZSQ and few-shot quantization\footnote{This refers to post-training quantization with few real data}(FSQ) commonly utilize FP32-pre-trained models (\textit{teacher}) to optimize quantized models (\textit{student}) by distilling knowledge.  
It is possible that ZSQ and FSQ share the quantization algorithm regardless of whether the data are real or synthetic. 
We thus adopt an up-to-date PTQ scheme to ZSQ so that breaking away from the quantization scheme conventionally used in ZSQ and then completing quantization within a few hours.
Based on the existing method, we propose a framework called~\genie\footnote{Data \underline{gen}eration scheme su\underline{i}t\underline{e}d for quantization} that distill data suited for model quantization. 
We also suggest a novel quantization scheme, a sub-module of \genie~that is available for both FSQ and ZSQ. As in Figure~\ref{fig:genie}, \genie~consists of two sub-modules: synthesizing data (\genie-D) and quantizing models (\genie-M).
By combining them, we bridge the gap between ZSQ and FSQ while taking an ultra-step forward from existing approaches. In other words, we achieve a state-of-the-art result that is unique among ZSQ approaches. 

Our contributions are summarized as follows:
\vspace{-2.5mm}
\begin{itemize}[leftmargin=1em]
    \item First, we propose a scheme for synthesizing datasets by combining the approaches related to generation and distillation to take advantage of both approaches.\vspace{-1.5mm}
    \item Second, we suggest a method to substitute convolution of stride $n$ ($n>1$) by \textit{swing convolution}. By applying randomness, various spatial information can be utilized when distilling datasets. \vspace{-1.5mm}
    \item Finally, we propose a new quantization scheme as a sub-module of \genie~(available for both FSQ and ZSQ), a simple but effective method that jointly optimizes quantization parameters.  
\end{itemize}

\section{Related Works}
\subsection{Uniform Quantization}

Uniform quantization maps full-precision weights into fixed-point numbers. Supposing the step size $s\in\mathbb{R}$ is set by a certain algorithm, we can obtain the integers of weights as follows:
\begin{align}
    \vw_{\text{int}}= clip \left(\left\lfloor\frac{\vw}{s}\right\rceil+\vz,n,p\right).
\end{align}
where $\lfloor\cdot\rceil$ denotes the nearest-rounding method, and $n$ and $p$ represent the lower and upper bound of the range, respectively. 
For example, when we asymmetrically quantize a layer to INT$b$, $n$ and $p$ are equal to $0$ and $2^{b-1}$, respectively. And the zero-point vector $\vz$ represents an all-$z$ vector, where $z=-\left\lfloor\frac{\min(\vw)}{s}\right\rceil$. 
Thus, the quantized weights $\vw^q$ can be represented as follows:
\begin{align}
    \vw^q=s(\vw_{\text{int}}-\vz).
\end{align}
To quantize a model, \textit{Min-Max} algorithm sets the step size $s$ as
\begin{align}
    s=\frac{\max(\vw)-\min(\vw)}{2^b-1},
\end{align}
where $b$ is the bit width for quantization. During optimization, \textit{Min-Max} updates $s$ in every step using an \textit{exponential moving average} (EMA) of $\min(\vw)$ and $\max(\vw)$, and update $\vw$ by \textit{straight-through estimator} (STE) (\ie, $\frac{\partial \gL}{\partial \vw}=\frac{\partial \gL}{\partial \vw^q}$)~\citep{bengio2013estimating}.
\textit{Learned step size quantization} (LSQ)~\citep{esser2019learned} learns the step size $s$ along with $\vw$ using STE, which is a state-of-the-art method in QAT.
Both algorithms are mainly used for net-wise optimization or QAT but can be used in a divide-and-conquer approach or in PTQ. 

\begin{figure*}[th]
    \centering
    \includegraphics[width=.83\linewidth]{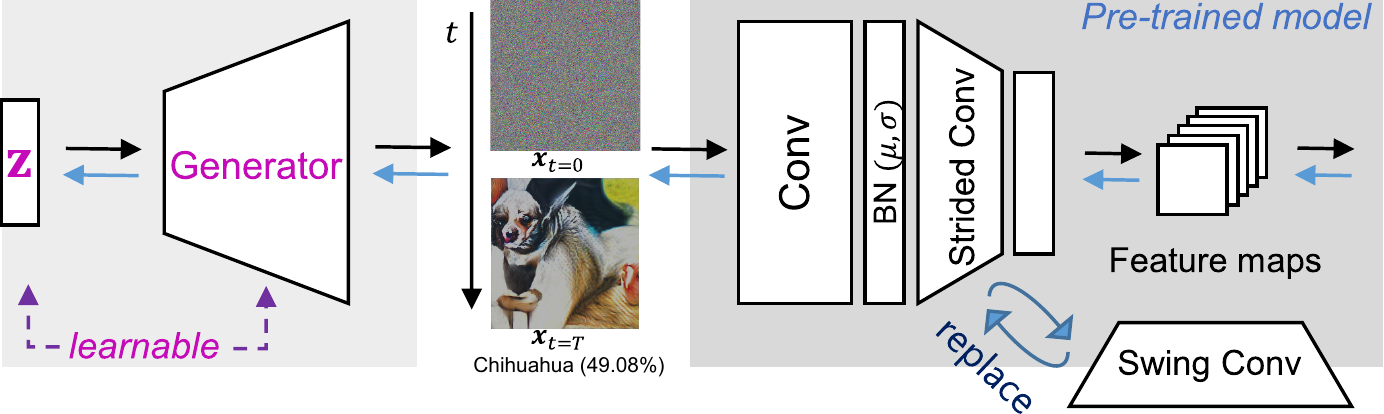}
    \caption{Data distillation (from noise ($t=0$) to signal ($t=T$)). We train the generator and latent vectors $\vz$, each of which is a kind of seed that generates its own image. The synthetic dataset is distilled indirectly by learning the latent vectors and the generator. When distilling images, $n(>1)$-stride convolutions on pre-trained models are replaced by \textit{swing convolutions}.}
    \label{fig:new}
\end{figure*}

PTQ~\citep{nagel2020up,li2021brecq,hubara2021accurate, jeon2022mr} optimizes networks mainly by the divide-and-conquer approach. Because they assume few shots are provided, they exploit knowledge from pre-trained models when quantizing models. 
AdaRound~\citep{nagel2020up} have suggested a rounding scheme while minimizing the reconstruction error for each layer between the pre-trained and quantized models (\ie, the mean squared error between activations of the two models). \textsc{Brecq}~\citep{li2021brecq} also has empirically shown the superiority of block-wise optimization. 
In contrast to ordinary knowledge distillation that aims to reduce the scale (\eg, the number of layers), the layers between \textit{teacher} and \textit{student} are mapped one-to-one in quantization, and thus models can be optimized per layer or per block in a divide-and-conquer approach. 
Although the latest algorithms for PTQ have been verified on real data, they can also be employed for ZSQ.

\subsection{Zero-shot Quantization}
Under the assumption that data are not available for quantization of models, ZSQ focuses on generating or distilling data by exploiting the information from pre-trained models. 
To synthesize the data, it uses the parameters in the batch normalization layers and assigns virtual hard labels $\vy$ to that synthetic data $\vx$ in order to utilize the \textit{CE loss}, which can be represented as follows:
\begin{align}
    \mathcal{L}_\mathrm{CE} = \mathbb{E} \left[\mathrm{CE}(f_{p}(\vx),\vy)\right],
    \label{eq:CE}
\end{align}
where $f_{p}$ denotes a pre-trained model.
In addition, Qimera~\citep{choi2021qimera} attempted to make boundary-supporting samples when synthesizing data for ZSQ. ZAQ~\citep{liu2021zero} generates data that maximize the discrepancy ($L_1$-distance) between the activations of the two models (\ie, \textit{teacher} and \textit{student}) while using the loss function that minimizes the discrepancy for quantization. 
To quantize models, GDFQ~\citep{xu2020generative} define the distillation loss as the combination of the \textit{CE loss} and \textit{KL divergence}. AIT~\citep{choi2022s} utilizes \textit{KL}-only loss based on the observation that it has flatter minima than \textit{CE loss}. 
Most of these quantization techniques optimize generators and quantized networks alternately while employing \textit{Min-Max} algorithm to quantize models. 
Furthermore, DFQ~\citep{nagel2019data} and ZeroQ~\citep{cai2020zeroq} utilize the synthetic data primarily to set the step size of the activations without gradient-based optimization.

\begin{table}[tb]
\renewcommand{\arraystretch}{1.2}
\footnotesize
\centering
\caption{Categorization of quantization algorithms} 
\vspace{-0.25cm}
\begin{tabular}{|l|c|c|}
\hline
 & Divide-and-Conquer & Netwise\tabularnewline
\hline
\multirow{2}{*}{Real Data} & \multirow{2}{2.5cm}{AdaRound, \textsc{Brecq}, \\ AdaQuant} & \multirow{2}{2.5cm}{LSQ, \textit{Min-Max}}\tabularnewline
 &  & \tabularnewline
\cline{1-3}
\multirow{2}{*}{Synthetic Data} & \multirow{2}{2.5cm}{\genie, MixMix} & \multirow{2}{2.5cm}{GDFQ, AIT, Qimera, ZAQ, IntraQ}\tabularnewline
 &  & \tabularnewline
\cline{1-3}
\end{tabular}
\label{tab:zsq}
\end{table}

Table~\ref{tab:zsq} summarizes quantization algorithms for both real- and synthetic datasets. 
According to our experiments, there is a limitation to the diversification of samples by synthesis. Thus, it is more suitable for ZSQ to use PTQ algorithms than QAT which can be overfitted with a small amount of the data (or monotonous data).

\section{\textsc{Genie}}

\subsection{Data Distillation (\textbf{\genie}-D)}
To synthesize data, ZSQ commonly utilizes statistics (mean $\mu$ and standard deviation $\sigma$) in the batch normalization layers (BNS) of the pre-trained models as follows:
\begin{align}
    \mathcal{L}^\mathrm{D}_\mathrm{BNS} = \sum_{l=1}^{L}(\norm{\boldsymbol{\mu}_{l}^s-\boldsymbol{\mu}_{l}}^2 + \norm{\boldsymbol{\sigma}_{l}^s-\boldsymbol{\sigma}_{l}}^2)
    \label{eq:BNS}
\end{align}
where $\boldsymbol\mu^s_l$/$\boldsymbol\sigma^s_l$ and $\boldsymbol\mu_l$/$\boldsymbol\sigma_l$ represent the statistical parameters of the synthetic data and learned parameters in the $l$-th batch normalization layer, respectively. 
To minimize Eq.~(\ref{eq:BNS}), generator-based schemes optimize weights in the generator, while distill-based schemes propagate the error directly to the synthetic data. 

\begin{figure*}[t]
  \centering
  \begin{subfigure}{0.45\linewidth}
    \centering
    \includegraphics[width=.80\linewidth]{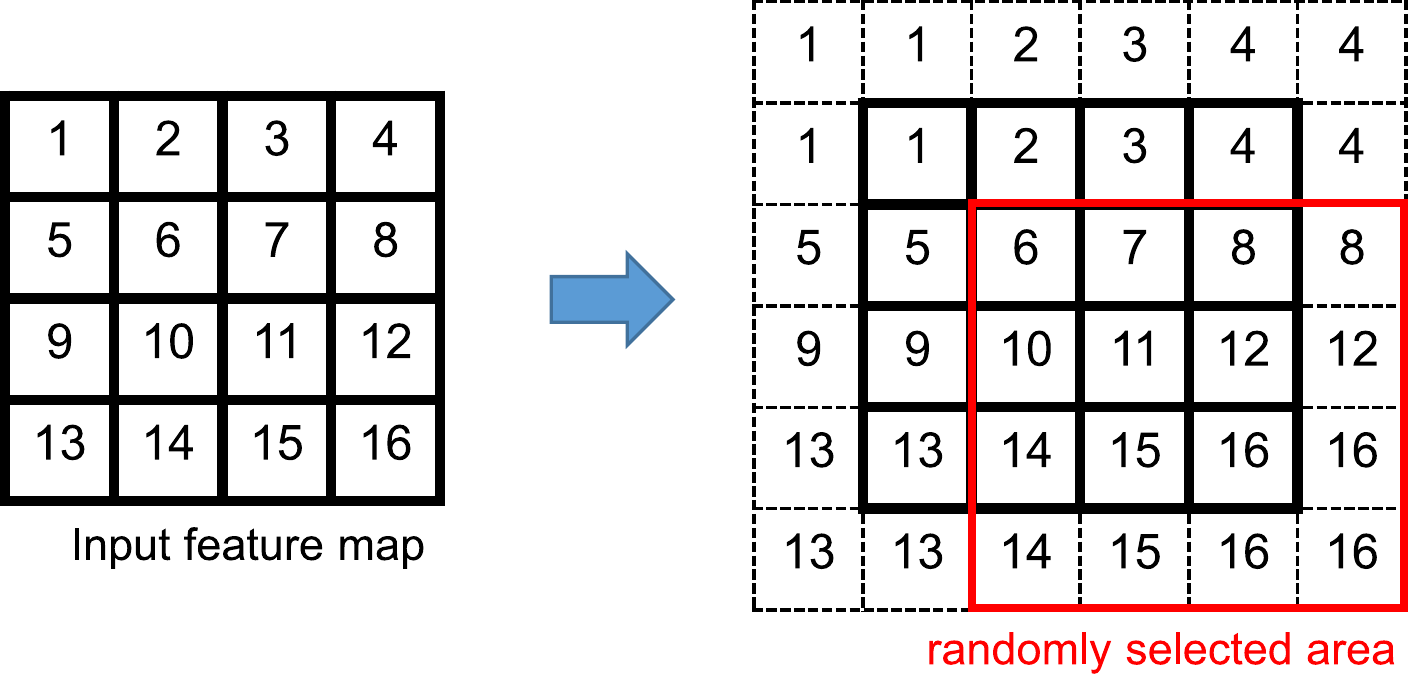}
    \caption{Reflection padding \& random crop.}
    \label{fig:swing}
  \end{subfigure}
  \hfill
  \begin{subfigure}{0.50\linewidth}
    \includegraphics[width=.97\linewidth]{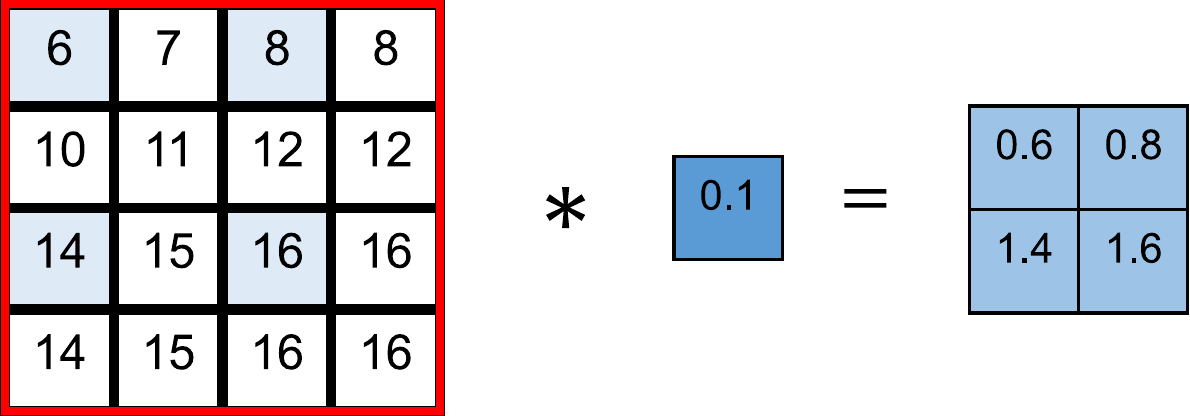}
    \caption{2-stride convolution (\texttt{conv2d(kernel\_size=1, stride=2)}).}
    \label{fig:sconv}
  \end{subfigure}
  \caption{\textit{Swing convolution}. (a) feature maps are extended by reflection padding and randomly cropped  (b) The randomly selected areas in feature maps are convolved with the stride of $n$ ($n>1$)).}
  \label{fig:swingconv}
\end{figure*}

Generator-based approaches (GBA)~\cite{choi2021qimera,choi2022s,xu2020generative,zhu2021autorecon,liu2021zero} use latent vectors from a Gaussian distribution ($\gN(0,\mI)$) as input of the generator in order to synthesize datasets. 
Thus, they have the advantage of synthesizing data infinitely as long as the input of the generator follows the designed distribution. 
Furthermore, the generator can be expected to learn \textit{common knowledge} of the input domain. 
However, GBAs have been attempting to optimize the generator such that converting all noise to semantic signals, which not only takes a long time to converge but also converges at a relatively high loss (\ie, a low
statistical similarity that is defined in Eq.~(\ref{eq:BNS})).
Although it is also possible to generate infinite data, the information required for quantization is redundant, which limits the enhancement of quantized networks, especially with QAT.

In contrast, distill-based approaches (DBA)~\cite{cai2020zeroq,zhong2022intraq,li2021mixmix} gradually update images from Gaussian random noises to semantic signals by distilling knowledge to the images. As it directly propagates the error to images, it converges relatively quickly to a low loss. However, there is no significant interaction between the instances except when measuring the loss in a batch. 

To take advantage of both approaches, inspired by \textit{Generative Latent Optimization} (GLO)~\citep{bojanowski2017optimizing}, we design a generator that produces synthetic data but distills the knowledge to latent vectors $\vz$ from a normal distribution. 
In other words, the synthetic images are distilled indirectly by the latent vectors which are trained and updated in every iteration.  
Figure~\ref{fig:new} illustrates the proposed method for distilling datasets.
The latent vector initialized in the Gaussian form becomes an image via the generator, and the image takes the loss from the pre-trained model; the latent vector and generator are updated by the loss.
The images from the initial vectors are close to noise, but they gradually mature into information suited for quantization as the optimization goes on. 
The image indicated by $\vx_{t=T}$ in Figure~\ref{fig:new} is a distilled image updated by the BNS loss (Eq.~\ref{eq:BNS}) and \textit{swing convolution} (Figure~\ref{fig:swingconv}) without any image prior loss. Genie-D synthesizes images that are very similar to the actual structure (Figure~\ref{fig:fig1} and~\ref{fig:new}).
By distilling latent vectors through the generator, \genie-D converges as fast as DBA while learning the \textit{common knowledge} of the input domain similar to GBA.

\revise{Indeed, we can consider GBA as Variational Auto-Encoder (VAE)~\cite{kingma2013auto} without the encoder. Suppose $\vx$ and $\vx'$ denote a real and fake sample (the output of the decoder or generator), respectively.
The generator can be optimized by minimizing the distance between $\vx$ and $\vx'$; it does not work well without the encoder that approximates true posterior $p_\theta(\vz|\vx)$. 
Without variational inference, it is trying to match $\vx'$ to $\vx$ in pixel space (which has high dimensional).
The distance loss can be replaced with BNS loss in GBA. 
Owing to the absence of real samples, GBA computes the distance indirectly by matching the distribution of the fake to its real (\ie, BNS loss) using pre-trained models.
However, GLO explains that generative models can be trained without the encoder by optimizing the latent vector~\cite{bojanowski2017optimizing}. 
Furthermore, we can efficiently explore attributes of the real images pre-trained models utilized by optimizing in manifold space (latent vector) ($n$-dimension) rather than optimizing in pixel space ($m$-dim, $n\ll m$), while training the generator for \textit{common knowledge} or image prior~\cite{jeon2021gradient,bau2019semantic,ulyanov2018deep}.
Which may explain the efficacy of optimizing latent vectors in addition to the generator compared to DBA and GBA.}

\newcommand{\tconv}{\textit{tconv}}
\newcommand{\stconv}{\textit{s-tconv}}
\newcommand{\conv}{\textit{conv}}
\newcommand{\sconv}{\textit{s-conv}}
\newcommand{\back}{\textit{backprop-op}}

\subsubsection{Swing Convolution}

When distilling images without a generator (\ie DBA), we can consider the back-propagating error (with respect to $\bns$) into the images as the process of image generation (\ie, model inversion~\cite{mordvintsev2015deepdream}). Moreover, the backpropagation function (\back) of convolutional layers (\conv) is transposed convolution (\textit{tconv}); the \back~for convolution of the stride $n$ ($n>1$, \sconv) is also $n$-stride \tconv~(\stconv). 
Because \stconv~(which is commonly used to increase the resolution of images) can create checkerboard artifacts when generating images~\cite{odena2016deconvolution}, 
the distilled images produced by \back~for \sconv~(\ie \stconv) during model inversion also can result in checkerboard artifacts, which degrade the quality of images owing to information loss.
Thus, we introduce \textit{swing convolution} performing stochastic $n$-stride \conv, which is simple but effective in reducing checkerboard artifacts caused by information loss and requires negligible extra computational cost.

\begin{figure}
  \centering
  \begin{subfigure}{0.45\linewidth}
    \centering
    \includegraphics[width=.95\linewidth]{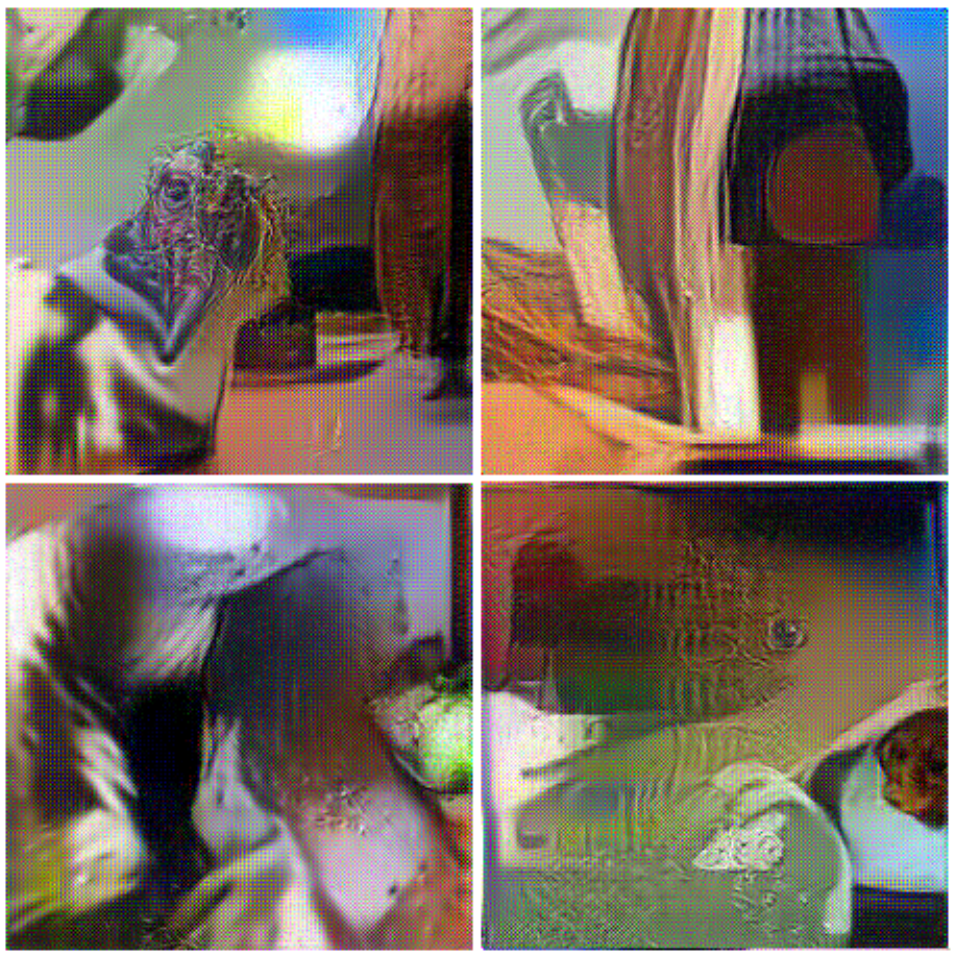}
    \caption{Distilling without \textit{swing conv}}
    \label{fig:woswing}
  \end{subfigure}
  \begin{subfigure}{0.45\linewidth}
    \centering
    \includegraphics[width=.95\linewidth]{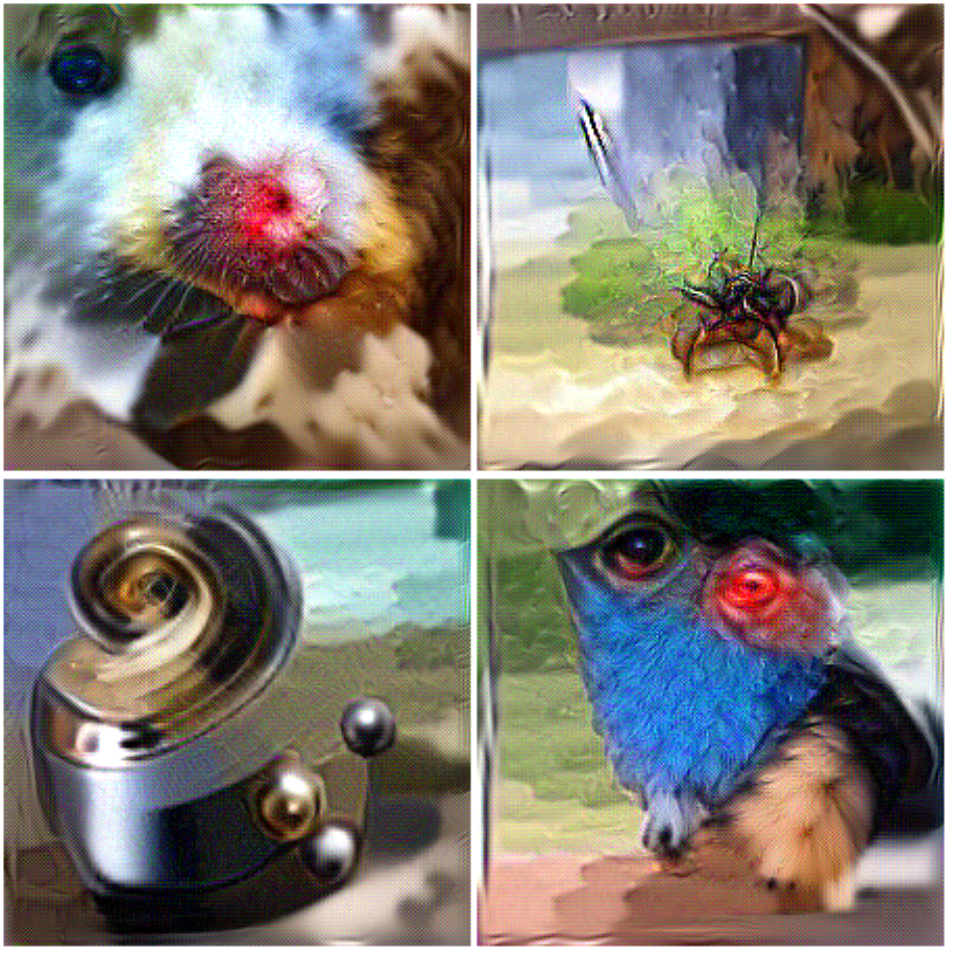}
    \caption{Distilling with \textit{swing conv}}
    \label{fig:wswing}
  \end{subfigure}
  \caption{The effect of \textit{swing convolution} that alleviates checkerboard artifacts resulting from information loss. The images in the same cell in each grid were distilled from the same seed. The images were directly distilled without the generator.}
  \label{fig:versus}
\end{figure}

With the reduction of information loss, distilled images become considerably robust and enhance the quality of quantized models. 
Figure~\ref{fig:swingconv} illustrates the mechanism of \textit{swing conv}. 
Before 2-stride \conv, the feature maps are extended by padding with their edge values (\ie, reflection padding) and randomly cropped to restore them to their original sizes, as shown in Figure~\ref{fig:swing}. 
Then, the randomly selected areas in the feature maps are stride-convolved as shown in Figure~\ref{fig:sconv}.
We refer to this series of processes as \textit{swing conv}. We replace all \textit{s-convs} to \textit{swing convs} when only synthesizing the datasets.
By applying randomness to the feature maps to be convolved, the distilled images can be updated so that the statistics of the outputs match the BNS regardless of the feature maps selected randomly.

Suppose that there is a $1\times1$ convolutional layer of stride $2$ with $4\times 4$ feature maps as input, pixels in the second and fourth row and columns are not utilized in the BNS loss and thus not used for back-propagation. With \textit{swing convolution}, however, all pixels in the feature maps for 2-stride \conv~can contribute toward distilling images across the optimization, which provides various spatial information with distilled images and results in enhancing the quantized models without information loss.
Note that shift operations such as \textit{swing conv} have been used to enhance models in various variations~\cite{zhao2017random,chen2019all,wang2022shift}. 
Especially, we have found that the mechanism of random shifting convolution~\cite{zhao2017random} is the same as \textit{swing convolution} although the purpose and way of use are different.
In model inversion, including distillation for ZSQ, to the best of our knowledge, this is the first instance of using stochastic convolution.

\begin{algorithm}[tb]
\footnotesize
\caption{Data distillation (\genie-D)} 
\label{algo:genieD}
\renewcommand\algorithmicrequire{\textbf{Input}:}
\renewcommand\algorithmicensure{\textbf{Output}:}
\begin{algorithmic}[1]
\Require Pre-trained model $f_p$
\Ensure Synthetic (distilled) data $\vx^r$
\Procedure{Data distillation}{$f_p$}
\State $\hat{f}_{p}=$ Strided\_Conv\_To\_Swing ($f_p$)
\State Init. latent vectors $\vz\sim\mathcal{N}(0,\mI)$ and weights $\mW_{\gG}$ of generator $\gG$
\Repeat
    \State $\vx^r=\gG(\vz)$
    \State Compute $\hat{f}_p(\vx^r)$
    \State Update $\vz$ and $\mW_\gG$ with respect to $\bns$\Comment{See Eq.(\ref{eq:BNS})}
\Until{converged}
\EndProcedure
\end{algorithmic}
\end{algorithm}

Algorithm~\ref{algo:genieD} describes our proposed method \genie-D used for synthesizing datasets. 
Before optimization, \genie-D replaces all \textit{sconvs} to \textit{swing convs} in the pre-trained model (\textit{line 2}). 
Note that the \textit{sconvs} are substituted by \textit{swing convs} only when distilling data and not during the quantization of models. 
After initializing the latent vectors $\vz$ and weights of the generator (\textit{line 3}), \genie-D optimizes them with respect to $\bns$ (\textit{line 4--8}). 
\revise{Moreover, we applied various methods including existing works to explicitly generate diverse data by modifying or adding any loss (including \textit{CE} loss) on top of \genie, but there was no significant improvement at least in PTQ.}

\subsection{Quantization Algorithm (\textbf{\genie}-M)}
\newcommand{\mma}{\textbf{M1}}
\newcommand{\mmb}{\textbf{M3}}
\newcommand{\mmc}{\textbf{M5}}
\newcommand{\mmd}{\textbf{M6}}
\newcommand{\mme}{\textbf{M7}}

\newcommand{\mmf}{\textbf{M2}}
\newcommand{\mmg}{\textbf{M4}}

\begin{table*}[tb]
\renewcommand{\arraystretch}{1.2}
\footnotesize
\centering
\caption{Result of the ablation study on CNN Models (top-1 accuracy (\%))}
\vspace{-0.25cm}
\begin{tabular}{l|c|cccc|rrrrr}
\hline
& \#Bits & \multicolumn{4}{c|}{Ablation Settings} &
\multirow{2}{*}{ResNet-18} &
\multirow{2}{*}{ResNet-50} &
\multirow{2}{*}{MobileNetV2} & 
\multirow{2}{*}{MobileNet-b} & 
\multirow{2}{*}{MnasNet-1.0} \\\cline{3-6}
  & (W/A) & Swing & Generator & $\vz$ & Genie-M &  &  & & &   \\ \hline
FP & 32/32 & & & & & {71.08} & {77.00} & {72.49} & {74.53} & {73.52}\\ \hline
\mma & \multirow{7}{*}{4/4} & & & & & 69.19 & 74.87 & \revise{66.22} & 66.01 & 58.52 \\
\revise{\mmf} & & & & & \checkmark & \revise{69.25} & \revise{74.94} & \revise{66.25} & \revise{66.45} & \revise{58.82} \\
\mmb & & \checkmark & & & & 69.49 & 75.43 &  67.80 & 67.14 & 63.98 \\
\revise{\mmg} & & & \revise{\checkmark} & & & \revise{69.17} & \revise{74.96} & \revise{66.41} & \revise{65.75} & \revise{64.63} \\
\mmc & & & \checkmark & \checkmark & & 69.58 & 75.39 & 67.92 &  67.40 & 66.15  \\
\mmd & & \checkmark & \checkmark & \checkmark & & 69.62 & 75.47 & 68.28 &  68.02 & 66.55 \\
\mme & & \checkmark & \checkmark & \checkmark & \checkmark &\textbf{69.66} & \textbf{75.59} & \textbf{68.38} &  \textbf{68.58} & \textbf{66.94} \\
\hline
\mma & \multirow{7}{*}{2/4} & & & & & 61.96 & 66.72 & \revise{36.58} & 23.18 & 31.22 \\
\revise{\mmf} & & & & & \checkmark & \revise{62.62} & \revise{66.95} & \revise{37.12} & \revise{27.44} & \revise{32.45} \\
\mmb & & \checkmark & & & & 63.74 & 69.44 &  44.00 & 27.85 & 34.64 \\
\revise{\mmg} & & & \revise{\checkmark} & & & \revise{60.13} & \revise{65.28} & \revise{34.92} & \revise{27.51} & \revise{35.50} \\
\mmc & & & \checkmark & \checkmark & & 64.06 & 70.16 & 47.96 & 32.53 & 45.47 \\
\mmd & & \checkmark & \checkmark & \checkmark  & & 64.34 & 69.87 & 49.89 & 36.48 & 47.34 \\
\mme & & \checkmark & \checkmark & \checkmark & \checkmark & \textbf{65.10} & \textbf{69.99} & \textbf{53.38} &  \textbf{48.36} & \textbf{48.21} \\
\hline
\end{tabular}
\label{tab:ablation}
\end{table*}

When quantizing a model of weights $\mW\in\mathbb{R}^{m\times n}$ with a fixed-point representation (\eg, INT8) in a post-training approach, we can set the step size (or scaling factor) $s$ from the pre-trained model as follows:
\begin{align}
    s^*=\arg\min_s\norm{\mW-s\cdot clip\left(\left\lfloor\frac{\mW}{s}\right\rceil,n,p\right)}_F,
    \label{eq:step}
\end{align}
where $\lfloor\cdot\rceil$ denotes the nearest-rounding method, and $n$ and $p$ represent the lower and upper bounds of the range, respectively. For example, when we symmetrically quantize a model to INT$b$, $n$ and $p$ are equal to $2^{b-1}$ and $2^{b-1}\texttt{-}1$, respectively. 
With the step size $s$, $\mW_{\text{int}}$ and $\mW^q$ can be defined as follows:
\begin{align}
    &\mW_{\text{int}}:=clip\left(\left\lfloor\frac{\mW}{s}\right\rceil,n,p\right)\\
    &\mW^{q}:=s\cdot\mW_{\text{int}}.
\end{align}
Using the above formulation, we can quantize neural networks even in the absence of data.
To further optimize the networks with an unlabeled dataset of small size, AdaRound~\citep{nagel2020up} proposed a rounding scheme that allocated weights to one of the two nearest quantization points.
Let the \textit{base} integer matrix $\mB\in[n,p]^{m\times n}$ be defined as  
\begin{align}
    &\mB:=clip\left(\left\lfloor\frac{\mW}{s}\right\rfloor,n,p\right),
    \label{eq:base}
\end{align}
where $\left\lfloor\cdot\right\rfloor$ denotes the floor function. 
Then, AdaRound sets the quantized weights $\mW^q$ as 
\begin{align}
    \mW^q=s\cdot\left(\mB+\mV\right)
    \label{eq:wq}
\end{align}
where $\mV\in[0,1]^{m\times n}$ is the \textit{softbit}\footnote{We have omitted details for a concise description. Refer to the appendix for further details on \textit{softbit}.} matrix to be optimized such that \revise{each weight $w_{\text{int},i}~(\in\mW_{\text{int}})$ is converged to either $b_i~(\in\mB)$ or $b_i+1$}.
However, AdaRound does not jointly optimize the step size $s$ with \textit{softbit} $\mV$, and the reason is explained as follows: "It is non-trivial to combine the two tasks: any change in the step size would result in a different quadratic unconstrained binary optimization (QUBO) problem"~\citep{nagel2020up}.
Because optimizing the step size $s$ results in the change in the \textit{base} $\mB$ (Eq.~(\ref{eq:base})), it can cause a conflict with the \textit{softbits} $\mV$ being optimized.
To resolve this issue, we suggest a method to enable joint optimization without conflict via a sub-module of \genie~(\genie-M), which is simple yet effective. 
Regardless of the $s$ being optimized, we maintain $\mB$ as initialized by releasing the mutual dependency between $\mB$ and $s$ (\ie, \texttt{$\mB$.detach()}). \revise{In other words, we consider $s$ as a learnable parameter that does not affect $\mB$.}
In Eq.~(\ref{eq:wq}), $\mB$ is considered constant and not dependent on $s$, and the losses propagated to $s$, $v~(\in\mV)$, and $b~(\in\mB)$ during optimization are computed as follows:
\begin{align}
\frac{\partial w^q}{\partial s}=b+v, ~~~ \frac{\partial w^q}{\partial v}=s ~~~\text{and}~~~
\frac{\partial w^q}{\partial b}=0.
\end{align}

\begin{algorithm}[tb]
\footnotesize
\caption{CLASS \genie-M}
\label{algo:mrbiq}
\renewcommand\algorithmicrequire{\textbf{class}:}
\renewcommand\algorithmicensure{\textbf{Output}:}
\renewcommand\algorithmicprocedure{\textbf{def}:}
\begin{algorithmic}[1]
\Procedure{$\_\_$init$\_\_$}{self, $\mW$, $bits$}
    \State self.$s~\gets$ SetStepSize($\mW$, $bits$) \Comment{Eq.~(\ref{eq:step})}
    \State self.$\mB~\gets$ clip$(\left\lfloor\frac{\mW}{\text{self.}s}\right\rfloor, n, p)$.detach() \Comment{Eq.~(\ref{eq:base})}
    \State self.$\mV~\gets$ $\frac{\mW}{self.s}-self.\mB$
\EndProcedure
\vspace{0.2cm}
\Procedure{forward}{self}
\State return self.$s\times$(self.$\mB$+self.$\mV$) \Comment{Eq.~(\ref{eq:wq})}
\EndProcedure
\end{algorithmic}
\end{algorithm}

Note that, such a joint optimization method can be applied to other post-training quantization algorithms such as AdaQuant~\cite{hubara2021accurate} in addition to AdaRound~\cite{nagel2020up}, that optimize only the integers of weights while maintaining the step size as the initialized states.

Algorithm~\ref{algo:mrbiq} presents the pseudo-code for \genie-M class, which is the sub-module used for distilling models in~\genie. Using this quantizer, we optimize the quantized models by minimizing the block-wise reconstruction error, like that in \textsc{Brecq}~\citep{li2021brecq}. \genie-M also uses LSQ~\citep{esser2019learned} to optimize the step size of activations, which can be combined with \textsc{QDrop}~\citep{wei2021qdrop}. All activations in a block are simultaneously optimized with all the weights in the block.

\begin{table*}[tb]
\renewcommand{\arraystretch}{1.2}
\footnotesize
\centering
\caption{Evaluation of CNN Models I (top-1 accuracy (\%))}
\begin{threeparttable}\vspace{-0.25cm}
\begin{tabular}{rlcrrrrr}
\hline
\multicolumn{2}{c}{Methods}      & \begin{tabular}[c]{@{}c@{}}\#Bits\\ (W/A)\end{tabular} & \multicolumn{1}{c}{ResNet-18} & \multicolumn{1}{c}{ResNet-50} & \multicolumn{1}{c}{MobileNetV2} & \multicolumn{1}{c}{MobileNet-b} & \multicolumn{1}{c}{MnasNet-1.0}  \\ \hline
\multicolumn{2}{c}{Full Prec.}   & 32/32   & {71.08}    & {77.00} & {72.49} & {74.53} & {73.52}\\ \hline
\multirow{6}{*}{\begin{sideways}Single Model\end{sideways}}   & ZeroQ+\textsc{Brecq}$^\ddagger$  & \multirow{11}{*}{4/4}  & 69.32 & 73.73 & 49.83 & 55.93 & 52.04\\
  & KW+\textsc{Brecq}$^\ddagger$  &   & 69.08 & 74.05 & 59.81 & 61.94 & 55.48\\
 & IntraQ$^\dagger$+\textsc{Brecq} &   & 68.77 & 68.16 & 63.78  & - & - \\
 & Qimera+\textsc{Brecq} &   & 67.86 & 72.90 & 58.33  & - & - \\
  & \textbf{\genie-D}+\textsc{Brecq} \textbf{[ours]} &   & {69.70} & 74.89 & 64.68 & {68.61} & 55.42 \\
  & \textbf{\genie~[ours]}   & &\textbf{69.66} & \textbf{75.59} & \textbf{68.38} & \textbf{68.58} & \textbf{66.94} \\
\cdashline{1-2}\cdashline{4-8}
\multirow{3}{*}{\begin{sideways} Mix* \end{sideways}} 
  & MixMix+\textsc{Brecq}$^\ddagger$    &   & 69.46 & 74.58 & 64.01 & 65.38 & 57.87  \\
  & \textbf{\genie-D}+\textsc{Brecq} \textbf{[ours]} &   & 69.71 & 74.89 & 64.97 & 62.70 & 51.25 \\
  & \textbf{\genie~[ours]}   &   & \textbf{69.77} & \textbf{75.41} & \textbf{68.70} & \textbf{69.04} & \textbf{67.45} \\
\cdashline{1-2}\cdashline{4-8}
\multirow{2}{*}{\begin{sideways}Real\end{sideways}}
& \textsc{QDrop}$^\S$   &  & 69.62 & 75.45 & 68.84 & - & -\\
& \textbf{\genie\revise{-M}~[ours]}   &   & \textbf{69.81}  & \textbf{75.61} & \textbf{69.23} & \textbf{69.80} & \textbf{68.29}\\
\hline
\multirow{6}{*}{\begin{sideways}Single Model\end{sideways}} 
    &   ZeroQ+\textsc{Brecq} & \multirow{11}{*}{2/4} & 61.63 & 64.16$^\ddagger$ & 34.39 & 23.53 & 13.83 \\
  & KW+\textsc{Brecq}$^\ddagger$   &  & - & 57.74 & - & - & -\\
 & IntraQ$^\dagger$+\textsc{Brecq} &   & 55.39 & 44.78 & 35.38  & - & - \\
   & Qimera+\textsc{Brecq} &   & 47.80 & 49.13 & 3.73  & - & - \\
   & \textbf{\genie-D}+\textsc{Brecq} \textbf{[ours]} &   & 64.24 & 69.38 & 45.28  & 42.50 & 29.72 \\
  & \textbf{\genie~[ours]}   &   & \textbf{65.10} & \textbf{69.99} & \textbf{{53.38}} & \textbf{48.36} & \textbf{48.21} \\
\cdashline{1-2}\cdashline{4-8}
\multirow{3}{*}{\begin{sideways} Mix* \end{sideways}} 
&MixMix+\textsc{Brecq}$^\ddagger$  &   & - & 66.49 & - & - & -  \\
& \textbf{\genie-D}+\textsc{Brecq} \textbf{[ours]} &   & 64.91 & 69.96 & 42.19 & 28.50 & 31.22 \\
& \textbf{\genie~[ours]}   &   & \textbf{65.44} & \textbf{70.62} & \textbf{53.36} & \textbf{49.89} & \textbf{49.65} \\
\cdashline{1-2}\cdashline{4-8}
\multirow{2}{*}{\begin{sideways}Real\end{sideways}}
& \textsc{QDrop}$^\S$   &  & 65.25 & 70.65 & 54.22 & - & -\\
& \textbf{\genie\revise{-M}~[ours]}   &  & \textbf{66.23} & \textbf{71.06} & \textbf{57.74} & \textbf{51.90} & \textbf{55.57}\\
\hline
\end{tabular}
\begin{tablenotes}
\item[$\ddagger$, $\S$] The figures are taken from ~\cite{li2021mixmix}$^\ddagger$ and ~\cite{wei2021qdrop}$^\S$. $^\dagger$ It synthesizes $5K$ images while others synthesize only $1K$ images.
\item[$*$] The synthetic datasets are distilled from multiple models like ensemble learning~\cite{li2021mixmix}.
\end{tablenotes}
\end{threeparttable}
\label{tab:ptq}
\end{table*}

\section{Experimental Results}
We evaluate our proposed method by testing it on convolutional neural networks (CNNs), such as ResNet~\citep{he2016deep}, MobileNet~\citep{howard2017mobilenets,sandler2018mobilenetv2}, RegNet~\citep{radosavovic2020designing}, and MnasNet~\citep{tan2019mnasnet}, where only $1K$ synthetic images distilled by \genie-D are used.
To optimize the quantized model, \genie-M quantizes each channel with asymmetrical ranges, whereas it quantizes the activations per tensor with symmetrical ranges.
\genie-M also exploits LSQ~\citep{esser2019learned} with QDrop~\cite{wei2021qdrop} for activation quantization.

\subsection{Ablation Study}

We build a combination of existing methods, such as ZeroQ+\textsc{QDrop}, and use it as the baseline of our approach (labeled \mma~in Table~\ref{tab:ablation}). 
Subsequently, we conduct an ablation study on various combinations to justify our design and verify the performance of \genie. The results of the ablation study are presented in Table~\ref{tab:ablation}:\vspace{-1.8mm}
\begin{itemize}[leftmargin=1em]
    \item \revise{\mma~vs. \mmd~and \mmg~vs. \mmd~describe the performance of \genie-D, the data distiller, compared to the existing methods such as ZeroQ (\mma) and GBA (\mmg).}\vspace{-2.0mm}
    \item \revise{\mmg~vs. \mmc~explains the efficacy when optimizing latent vector in addition to the generator.}\vspace{-2.0mm}
    \item \mmb~and \mmc~describe the performance of each factor constitutive of \genie-D. In \mmc, images are distilled indirectly by training the latent vector $\vz$ without replacing the strided convolution with \textit{swing}, while images in \mmb~are distilled directly without the generator but with \textit{swing}.\vspace{-2.0mm}
    \item \revise{\mma~vs. \mmf~and} \mmd~vs. \mme~show the performance of \genie-M, the model distiller, compared to that of an existing PTQ scheme (\ie, \textsc{QDrop}~\cite{wei2021qdrop}). \vspace{-2.0mm}
    \item \mme~demonstrates the performance when harmonizing the proposed methods, \genie-M and \genie-D.
\end{itemize}

\subsection{Performance of \textbf{\textsc{Genie}}}

\begin{figure}[t]
  \centering
  \includegraphics[width=0.99\linewidth]{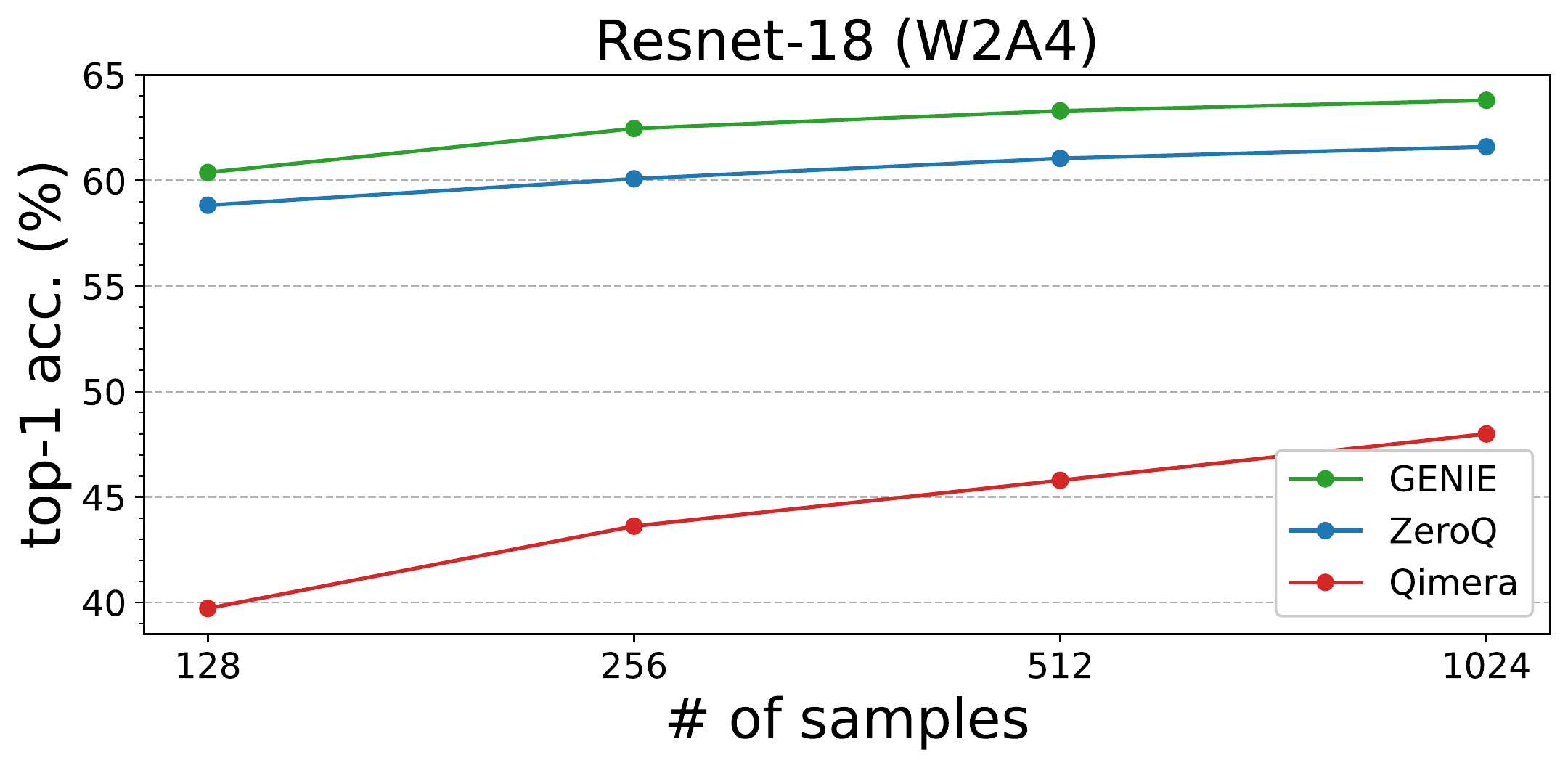}
  \caption{The influence of the number of samples on ResNet-18}
  \label{fig:numsample}
\end{figure}

To verify the performance of \genie, we compare various ZSQ algorithms, including ZeroQ~\citep{cai2020zeroq}, 
KW~\cite{haroush2020knowledge}, GDFQ~\citep{xu2020generative}, Qimera~\citep{choi2021qimera}, ARC~\citep{zhu2021autorecon}, AIT~\citep{choi2022s}, ZAQ~\citep{liu2021zero}, MixMix~\cite{li2021mixmix} and IntraQ~\citep{zhong2022intraq}.
Tables~\ref{tab:ptq} and~\ref{tab:cnn} summarize the comparison results.
In Table~\ref{tab:ptq}, we present the comparison of \genie~with other methods that utilize \textsc{Brecq} as the quantizer. Among them, MixMix distills images from various models (a total of twenty-one) like ensemble learning. 
To compare with MixMix, we also distilled images from five models of Table~\ref{tab:ptq}.
Despite ensembling fewer models, Genie-M achieves superiority over MixMix.
Using the same quantizer (\textsc{Brecq}), we verified the performance of \genie-D, the data synthesizer.
\revise{Even with 256 images, \genie~(62.46\%) also shows better performance compared to others with 1K images as shown in Figure~\ref{fig:numsample}: Qimera (47.99\%) and ZeroQ (61.6\%).}
\revise{Based on the observation, we can consider the fake data distilled by \genie-D more informative. The influence on other models can be found in the appendix.}
All methods in Table~\ref{tab:ptq} and Figure~\ref{fig:numsample} quantized the first and the last layer into 8 bits like that in \textsc{Brecq}. 
\begin{table}[tb]
\renewcommand{\arraystretch}{1.2}
\footnotesize
\centering
\caption{Evaluation of CNN Models II (top-1 accuracy (\%))}
\begin{threeparttable}\vspace{-0.25cm}
\begin{tabular}{lp{0.2cm}rrr}
\hline
\multicolumn{1}{c}{Methods}      & & \multicolumn{1}{c}{ResNet-18} & \multicolumn{1}{c}{ResNet-50} & \multicolumn{1}{c}{MobileNetV2} \\ \hline
\multicolumn{1}{c}{Full Prec.}   &  & {71.47}    & {77.73} & {73.03} \\ \hline
     GDFQ+AIT*  & \multirow{7}{*}{4/4} & 65.51 & 64.24 & 65.39   \\
     Qimera+AIT*  &   & 66.83 & 67.63 & 66.81   \\
     ARC+AIT*  &   & 65.73 & 68.27 & 66.47   \\
     ZAQ$\dagger$  &   & - & 70.06 & -   \\
     IntraQ$^\ddagger$  &   & 66.47 & - & 65.10 \\
     \revise{\textbf{\genie-D}+AIT}   &    & \revise{66.91} & - & -  \\ 
    \textbf{\genie~[ours]}   &    & \textbf{68.69} & \textbf{74.21} & \textbf{69.59}  \\ \hline
    GDFQ+AIT  & \multirow{7}{*}{2/4} & 0.10 & 0.10 & 0.11   \\
     Qimera+AIT  &   & 0.10 & 0.10 & 0.12   \\
     ARC+AIT  &   & 0.11 & 0.10 &  0.13  \\
     IntraQ  &   & 0.14 & - & 0.17 \\
     \revise{\textbf{\genie-D}+AIT}   &    & \revise{\textbf{0.50}} & - & -  \\ 
    \textbf{\genie~[ours]}   &    & \textbf{58.73} & \textbf{54.83} & \textbf{45.84}  \\ 
    \hline 
\end{tabular}
\begin{tablenotes}
\item[*,$\dagger\text{,}\ddagger$] The figures are taken from~\cite{choi2022s}$^*$,~\cite{liu2021zero}$^\dagger$, and ~\cite{zhong2022intraq}$^\ddagger$.
\end{tablenotes}
\end{threeparttable}
\label{tab:cnn}
\end{table}

\begin{table*}[th]
\renewcommand{\arraystretch}{1.2}
\footnotesize
\centering
\caption{Performance comparison using \textbf{\textit{real samples}} ($1K$) (top-1 Accuracy (\%))}
\begin{threeparttable}\vspace{-0.25cm}
\begin{tabular}{lcrrrrrr}
\hline
\multicolumn{1}{c}{Methods}      & \begin{tabular}[c]{@{}c@{}}\#Bits\\ (W/A)\end{tabular} & \multicolumn{1}{c}{ResNet-18} & \multicolumn{1}{c}{ResNet-50} & \multicolumn{1}{c}{MobileNetV2} & \multicolumn{1}{c}{RegNetX-600M} & \multicolumn{1}{c}{RegNetX-3.2G} & \multicolumn{1}{c}{MnasNet-2.0} \\ \hline
\multicolumn{1}{c}{Full Prec.}   & 32/32   & {71.08}    & {77.00} & {72.49} & {73.71} & {78.36} & {76.68} \\ \hline
    AdaRound+\textsc{QDrop}$^\dagger$  & \multirow{3}{*}{4/4}  & 69.10 & 75.03 & 67.89 & 70.62 & 76.33 & 72.39 \\
    \textbf{\genie-M}+No Drop \textbf{[ours]}     &   & 69.13  & 74.93  & 68.22  & 70.87  & 76.50 & 72.68  \\
    \textbf{\genie-M}+\textsc{QDrop} \textbf{[ours]}     &   & \textbf{69.35}  & \textbf{75.21} & \textbf{68.65}  & \textbf{71.13}  & \textbf{76.75} & \textbf{73.37}  \\
\hline 
    AdaRound+No Drop$^\dagger$  & \multirow{4}{*}{2/4}& 64.16 & 69.60 & 51.61 & 61.52 & 70.29 & 60.00 \\
    AdaRound+\textsc{QDrop}$^\dagger$  & & 64.66 & 70.08 & 52.92 & 63.10 & 70.95 & 62.36 \\
    \textbf{\genie-M}+No Drop \textbf{[ours]}    &   & 65.27  & 70.39  & 55.55  & 63.66  & 71.79 & 62.76  \\
    \textbf{\genie-M}+\textsc{QDrop} \textbf{[ours]}    &   & \textbf{65.77}  & \textbf{70.51}  & \textbf{56.38}  & \textbf{64.55}  & \textbf{72.35} & \textbf{64.10}  \\
\hline
    AdaRound+\textsc{QDrop}$^\dagger$  & \multirow{3}{*}{3/3} & 65.56 & 71.07 & 54.27 & 64.53 & 71.43 & 63.47 \\
    \textbf{\genie-M}+No Drop \textbf{[ours]}    &   & 65.50  & 71.08  & 55.28  & 64.37  & 72.05 & 62.17  \\
    \textbf{\genie-M}+\textsc{QDrop} \textbf{[ours]}     &   & \textbf{66.16}  & \textbf{71.61}  & \textbf{57.54}  & \textbf{65.68}  & \textbf{72.72} & \textbf{64.80}  \\
\hline
    AdaRound+No Drop$^\dagger$ & \multirow{4}{*}{2/2}  &46.64 & 47.90 & 4.55 & 25.52 & 39.76 & 9.51\\
    AdaRound+\textsc{QDrop}$^\dagger$  & & 51.14 & 54.74 & 8.46 & 38.90 & 52.36 & 22.70 \\
    \textbf{\genie-M}+No Drop \textbf{[ours]}     &   & 50.52  & 51.80  & 12.63  & 34.03  & 40.97 & 19.60  \\
    \textbf{\genie-M}+\textsc{QDrop} \textbf{[ours]}     &   & \textbf{53.71}  & \textbf{56.71}  & \textbf{17.10}  & \textbf{42.00}  & \textbf{55.31} & \textbf{28.56}  \\
\hline
\end{tabular}
\begin{tablenotes}
\item[$^\dagger$]The figures are taken from~\cite{wei2021qdrop}.
\end{tablenotes}
\end{threeparttable}
\label{tab:real}
\end{table*}

For a fair comparison, we also quantize the first and last layers and follow the setting of quantization points of GDFQ, AIT, ZAQ, and IntraQ. The results are presented in Table~\ref{tab:cnn}, where the pre-trained models we use are the same as they utilized.
As shown in the table, \genie~has significant differences from existing methods. 
Notably, when the bit width is W2A4, \genie~outperforms other methods, which empirically shows that algorithms for PTQ are more suitable for ZSQ than schemes for QAT. 
In Table~\ref{tab:gpuh}, we also measure the elapsed time to complete ZSQ on Nvidia V100. 
Because GBA such as GDFQ, Qimera, and AIT utilize QAT scheme for quantization, they devote quantization more time rather than training the generator. 
In contrast, \genie~focuses more on training the generator with the latent vectors while reducing the time to quantize models by using PTQ.
In summary, the PTQ approach with distilled data is very efficient for both time and accuracy in ZSQ.

\begin{table}[tb]
\renewcommand{\arraystretch}{1.2}
\scriptsize
\centering
\caption{Elapsed time to complete ZSQ (Hours)}
\begin{threeparttable}\vspace{-0.25cm}
\begin{tabular}{ccccc}
\hline
  \begin{tabular}[c]{@{}c@{}}\\\end{tabular}&\begin{tabular}[c]{@{}c@{}}GDFQ\\ +AIT\end{tabular} & \begin{tabular}[c]{@{}c@{}}Qimera\\ +AIT\end{tabular} & \begin{tabular}[c]{@{}c@{}}ARC\\ +AIT\end{tabular} &
  \multicolumn{1}{c}{\genie} \tabularnewline
\hline
\hline
ResNet-18 & 10.11 (1.71) & 10.19 (2.23) & 19.04 (10.56) &  2.73 (2.40) \\
ResNet-50 &  21.03 (2.71) & 19.08 (5.10) &  25.01 (10.96) &  6.22 (5.07) \\
MobileNetV2   & 17.21 (2.15)  &  19.50 (3.61) & 25.75 (10.26) & 4.10 (3.33) \\
\hline
\end{tabular}
\begin{tablenotes}
\item[*] The number in brackets denotes the elapsed time to train the generator.
\end{tablenotes}
\end{threeparttable}
\label{tab:gpuh}
\end{table}

\subsection{Comparison using Real Data}

We conduct experiments with randomly sampled datasets from ImageNet (ILSVRC12)~\citep{russakovsky2015imagenet} on various CNN models to verify our quantization algorithm, \genie-M.
Table~\ref{tab:real} shows the results of the comparisons; all methods quantized the first and the last layer into 8 bits like that in \textsc{QDrop}~\cite{wei2021qdrop}, a state-of-the-art method used for PTQ.
\textsc{QDrop} is a scheme for generalizing quantized models by randomly dropping the quantization operation for activations, such as Dropout~\cite{srivastava2014dropout} or DropConnect~\cite{wan2013regularization}. 
\textsc{QDrop} thus can be compatible with other quantization schemes such as LSQ. As shown in the table, \genie-M outperforms the existing methods while verifying the effect of joint optimization.
We use the average values in the table after evaluating the accuracy of 20 runs with randomly sampled images.

\section{Conclusion}
We have proposed a framework called \genie~for ZSQ. By distilling the latent vectors through the generator, \genie-D converges quickly to a low loss while learning \textit{common knowledge} of the input domain from the pre-trained model. 
By replacing the $n$-stride convolutions ($n\texttt{>}1$) with \textit{swing convolutions}, \genie~minimizes the information loss, resulting in enhanced performance of the quantized models.
We have also suggested a PTQ scheme that jointly optimizes quantization parameters regardless of the data properties. 
In summary, \genie~has achieved a new state-of-the-art performance on both ZSQ and FSQ.\\

\subsection*{Acknowledgements}
\noindent We would like to appreciate Kyungphil Park for his help in conducting the experiments and Seijoon Kim, Ph.D. for his helpful discussion.

{\small
\bibliographystyle{ieee_fullname}
\bibliography{99bib, 99cvpr}
}

\newpage
\clearpage
\appendix

\renewcommand{\thefigure}{A\arabic{figure}}
\renewcommand{\theequation}{A\arabic{equation}}
\renewcommand{\thetable}{A\arabic{table}}
\renewcommand{\thealgorithm}{A\arabic{algorithm}}
\setcounter{figure}{0}
\setcounter{equation}{0}
\setcounter{table}{0}
\setcounter{algorithm}{0}
\section{Implementation Details}

For all experiments related to data distillation, we set the batch size to 128 and used the Adam~\cite{kingma2015adam} optimizer with an initial learning rate of 0.01 and 0.1 for the generator and latent vectors, respectively. 
The learning rate for the generator decays exponentially by gamma (= 0.95) every 100 steps, whereas the learning rate for the latent vectors is scheduled with "\textit{ReduceLROnPlateau}" like that in ZeroQ~\cite{cai2020zeroq}. For all experiments, we distilled 1024 images, which were used for quantization. 
Each batch was independently distilled, and the weights of the generator were shared only within a batch.
In other words, the weights of the generator are initialized when distilling another batch.

For model distillation, we set the batch size to 32 and used the Adam optimizer to train the quantization parameters, namely, the scaling factor $s_w$, \textit{softbit} $\mV$, and step size $s_a$ of activations, the initial learning rates for which were 0.0001, 0.001, and 0.00004, respectively.
We also used cosine annealing~\cite{loshchilov2017sgdr} to decay the learning rate to 0 for the scaling factors of weights and step size of activations during optimization. 
We obtained pre-trained models from public repositories\footnote{https://github.com/yhhhli/BRECQ}\textsuperscript{,}\footnote{https://github.com/osmr/imgclsmob}.

\section{Block-Wise Optimization}

To optimize the quantized models, we minimize the reconstruction error between two blocks, which is sequentially performed from the input layers as follows:
\begin{align}
    \underset{s_w, s_a, \mV}{\arg\min}\norm{\vz-\vz^q}^2_2,
    \label{eq:brec}
\end{align}
where $\vz$ and $\vz^q$ are the outputs of the two blocks in the pre-trained \textit{teacher} and quantized \textit{student} models, respectively.
Subsequently, we ensure that the \textit{softbits} $h(\mV)$\footnote{For a brief explanation, we notated \textit{softbits} as only $\mV$ with the sigmoid function $h(\cdot)$ omitted, in the manuscript.} takes 0 or 1 by adding the regularization term to Eq.~(\ref{eq:brec}). Namely,
\begin{align}
    \underset{s_w, s_a,\mV}{\arg\min}\norm{\vz-{\vz}^q}^2_2+\lambda\sum_{i,j}(1-\big|2h(\mV_{i,j})-1\big|^\beta),
    \label{eq:obj}
\end{align}
where $h(\cdot)$ denotes the rectified sigmoid function~\cite{louizos2018learning} and $\beta$ is annealed during optimization like that in AdaRound~\cite{nagel2020up}. 
The Lagrange multiplier $\lambda$ in Eq.~(\ref{eq:obj}) is set to 1.0 or 0.1 for all experiments involving\ \genie-M, \textsc{QDrop}~\cite{wei2021qdrop} and \textsc{Brecq}~\cite{li2021brecq}, respectively.
Algorithm~\ref{algo:genie_m} summarizes our quantization approach, where we assume that a block consists of one layer for a concise explanation (\ie, layer-wise optimization). In practice, we designate a block (that consists of consecutive layers) as a residual block, like that in \textsc{QDrop} and \textsc{Brecq}.

\begin{algorithm}[t]
\footnotesize
\caption{Layer-wise reconstruction for quantization} 
\label{algo:genie_m}
\renewcommand\algorithmicrequire{\textbf{Input}:}
\renewcommand\algorithmicensure{\textbf{Output}:}
\begin{algorithmic}[1]
\Require Full-precision weights $\mW\in\mathbb{R}^{m\times n}$ and input activations $\vx$
\Ensure Quantized layer
\Procedure{Genie-M}{$\mW, \vx$, $bits$}
\State WeightQuant$\gets$\textsc{Genie}-M($\mW, bits$)\Comment{Algorithm \ref{algo:mrbiq}}
\State $s_a$=Initialize($\vx_1$) \Comment{Init. step size of act. by 1st batch}
\For {each input $\vx$} 
    \State $\vy=\mW\cdot\vx$
    \State $\vx^q \gets$ $s_a\cdot \left\lceil\frac{\vx}{s_a}\right\rfloor$\Comment{Act. quant. using LSQ~\cite{esser2019learned}}
    \State $\mW^q\gets$ WeightQuant( )
    \State $\hat{\vy}\gets\mW^q\cdot\vx^q$
    \State $\mathcal{L}\gets ||\hat{\vy}-\vy||^2_2+\lambda f_{reg}(\mV)$\Comment{See Eq.~(\ref{eq:obj})}
    \State $\mathcal{L}.$\texttt{backward()}\Comment{Update $s_w$, $s_a$, and $\mV$ with respect to $\mathcal{L}$}
\EndFor
\EndProcedure
\end{algorithmic}
\end{algorithm}

\begin{figure}
    \centering
    \includegraphics[width=1.0\linewidth]{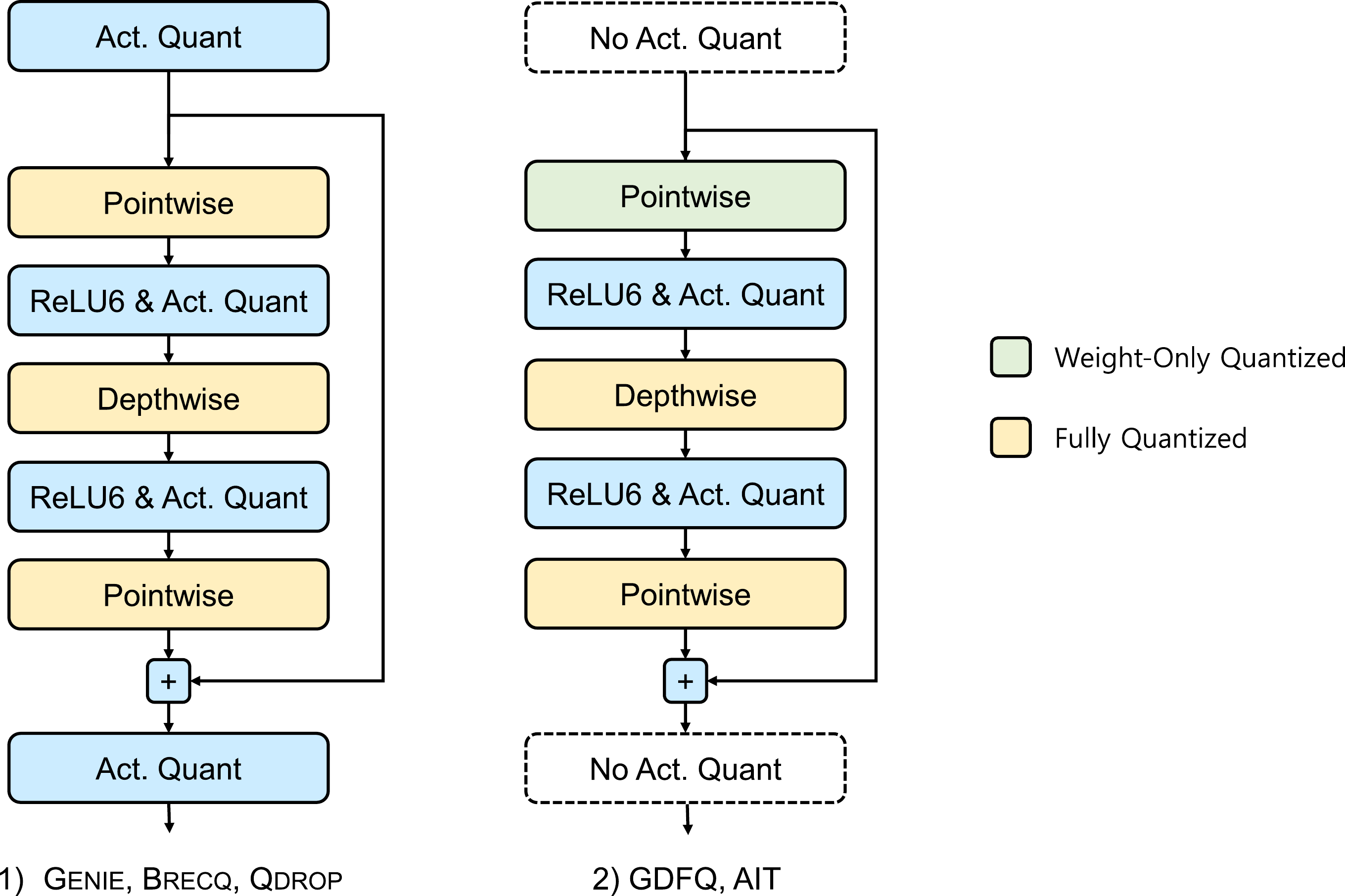}
    \caption{Comparison of quantization points in the inverted residual block of MobileNetV2, where AIT does not quantize the output of the block.}
    \label{fig:qpoints}
\end{figure}

\section{Quantization Setting}

When quantizing models into W$w$A$a$, \textsc{QDrop} quantizes the weights and activations into $w$ and $a$, respectively, excluding the first and last layers where the weights of the first and last and input activations of the last layers are quantized into 8-bit fixed-point numbers. \textsc{Brecq} additionally quantizes the output activations of the first layer into 8-bit while the other weights and activations are the same as in \textsc{QDrop}.
In contrast, AIT~\citep{choi2022s} quantizes the weights and activations of all layers (including the first and last layers) into $w$ and $a$, respectively.
AIT quantizes activations only after the activation functions, so quantizing activations is often omitted when there is no activation function at the end of the residual block, as in MobileNetV2~\cite{sandler2018mobilenetv2} and MnasNet~\cite{tan2019mnasnet} (Figure~\ref{fig:qpoints}). 
The methods depicted in Tables~\ref{tab:ablation} and~\ref{tab:ptq} follow the settings of \textsc{Brecq}, whereas those depicted in Tables~\ref{tab:cnn} and~\ref{tab:real} follow the setting of AIT and \textsc{QDrop}, respectively.

\section{Effect of the Initial Step Size}

\begin{figure}[t]
    \centering
    \includegraphics[width=1.0\linewidth]{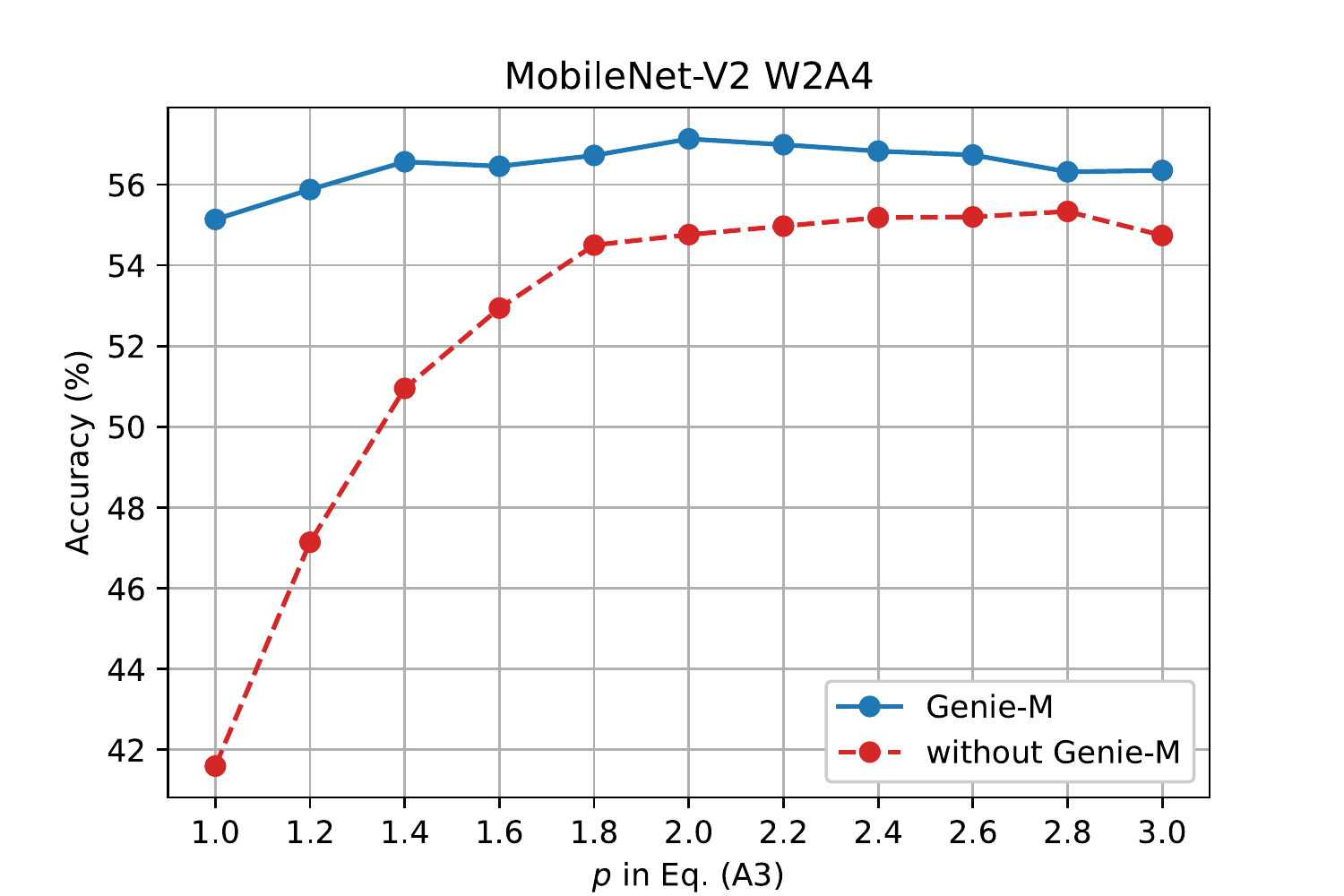}
    \caption{Comparison of accuracy depending on the $p$ in Eq.~(\ref{eq:step_init}).}
    \label{fig:init_norm_ablation}
\end{figure}

Figure~\ref{fig:init_norm_ablation} shows the performance of the quantized models depending on $p$ used when the step size of weights is initialized as follows:
\begin{align}
    s^*=\underset{s}{\arg\min}\norm{\mW-s\cdot clip\left(\left\lfloor\frac{\mW}{s}\right\rceil,\text{n}, \text{p}\right)}_{p,p}.
    \label{eq:step_init}
\end{align}
Because \textsc{QDrop} and \textsc{Brecq} maintain the initial step size of weights during optimization, the initialized step size can affect the performance of the quantized models. In contrast, \genie-M learns the step size, and thus the initial step size has a negligible impact on the accuracy.

\begin{table*}[t]
\renewcommand{\arraystretch}{1.2}
\footnotesize
\centering
\caption{The influence of the number of samples on model accuracy (W2A4)}
\begin{tabular}{c|ccc|ccc|ccc}
\hline
 & \multicolumn{3}{c|}{ResNet-18} & \multicolumn{3}{c|}{ResNet-50} & \multicolumn{3}{c}{MobileNetV2} \\ \hline
\# samples & ZeroQ    & Qimera    & \genie   & ZeroQ    & Qimera    & \genie   & ZeroQ     & Qimera    & \genie    \\ \hline
128        & 57.99    & 38.29     & 60.50   & 61.98    & 27.25     & 65.04   & 30.59     & 0.97     & 44.57    \\
256        & 59.67    & 42.82     & 62.26   & 64.21    & 33.23     & 67.23   & 34.25     & 1.22     & 47.66    \\
512        & 61.00    & 45.92     & 63.39   & 65.76    & 38.10     & 68.51   & 35.71     & 1.38     & 49.69    \\
1024       & 61.96    & 47.96     & 64.34   & 66.72    & 41.00     & 69.87   & 36.58     & 1.40     & 49.89    \\ 
2048       & 62.68    & 48.79     & 64.72   & 67.37    & 43.85     & 69.99   & 36.74     & 1.40     & 50.78    \\ \hline
\end{tabular}
\label{tab:num_sample}
\end{table*}
\begin{figure}
    \centering
    \includegraphics[width=0.9\linewidth]{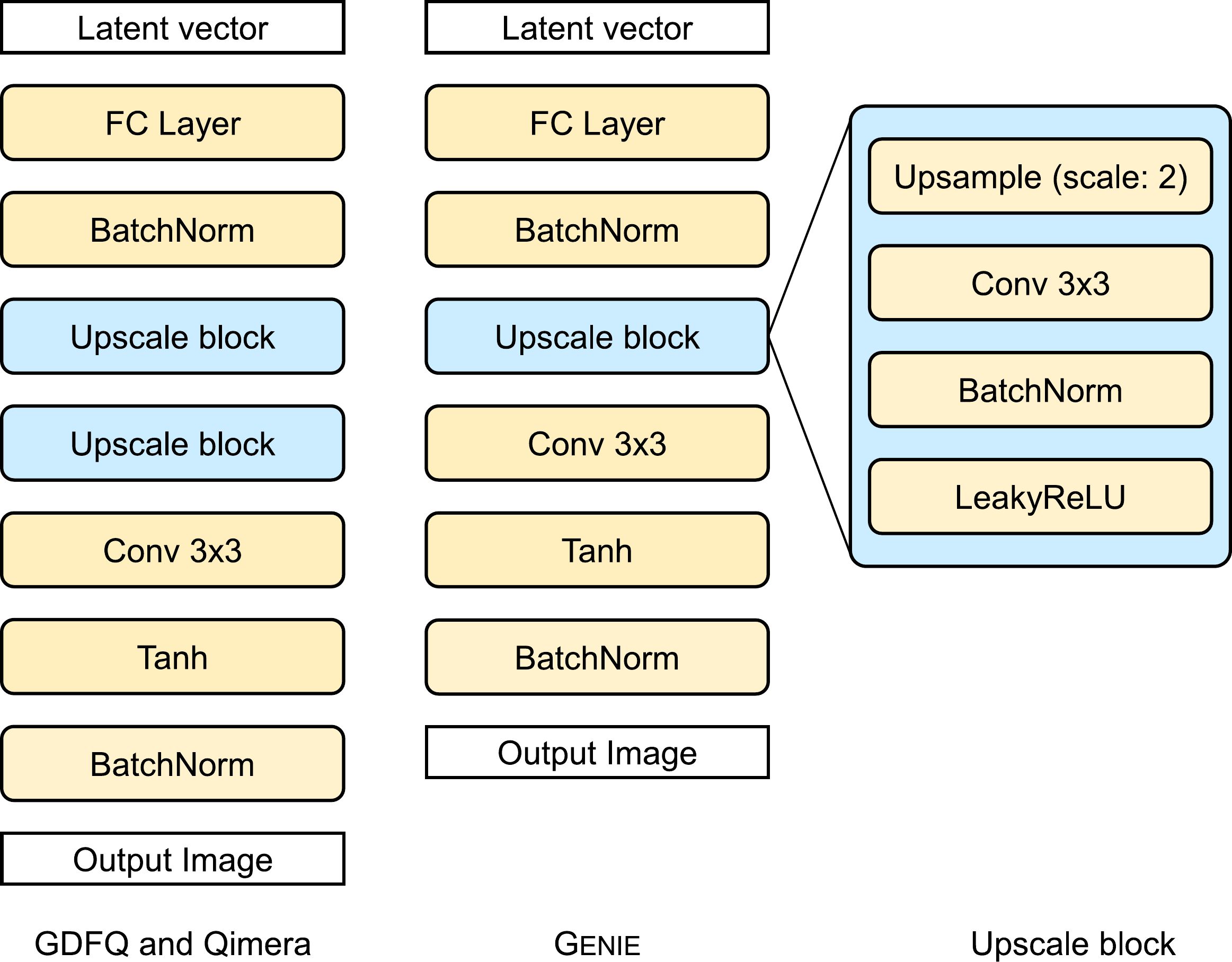}
    \caption{Structure of the generator and upscale block}
    \label{fig:gen}
\end{figure}

\begin{figure}[t]
  \centering
  \begin{subfigure}{0.95\linewidth}
    \centering
    \includegraphics[width=\linewidth]{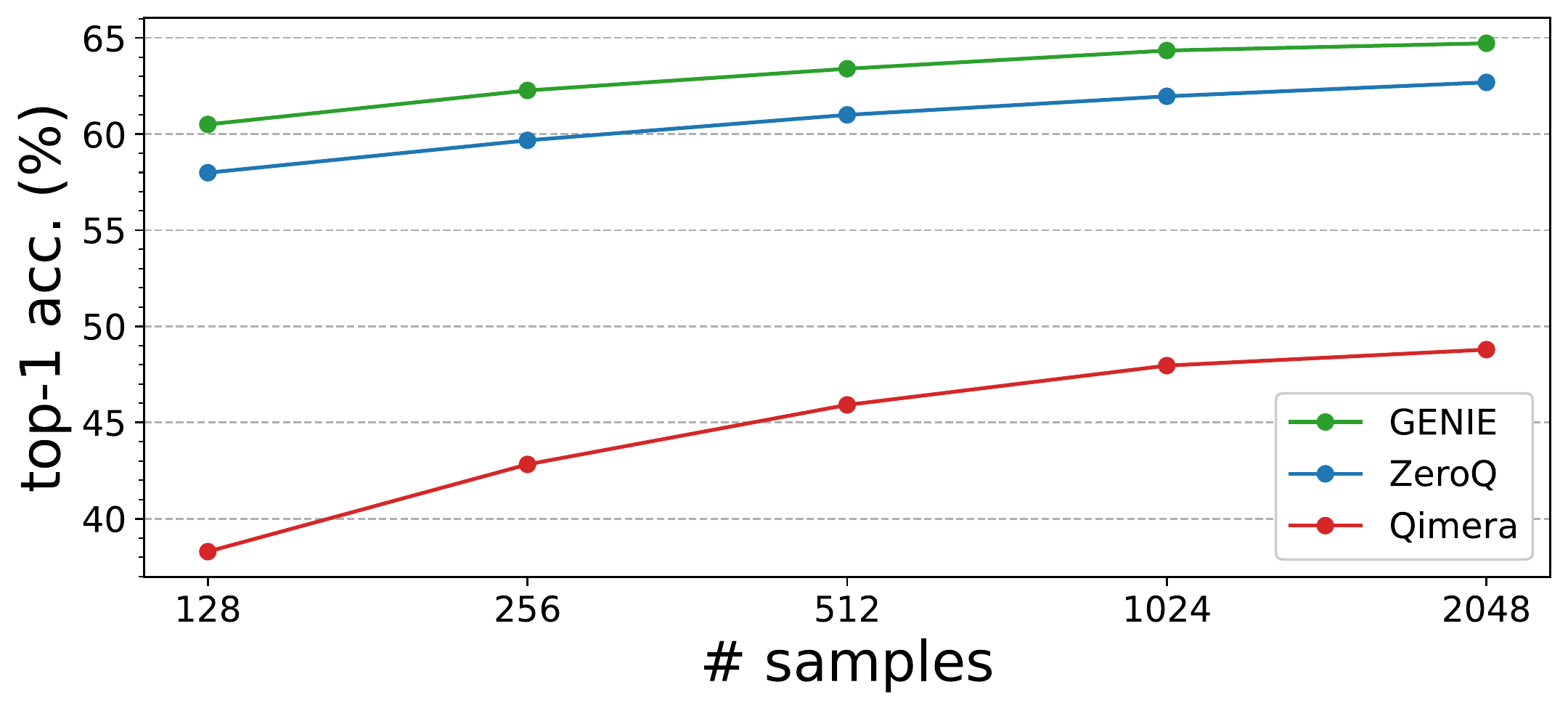}
    \caption{ResNet-18}
    \label{fig:num_a}
  \end{subfigure}
  \hfill
  \begin{subfigure}{0.95\linewidth}
    \centering
    \includegraphics[width=1.0\linewidth]{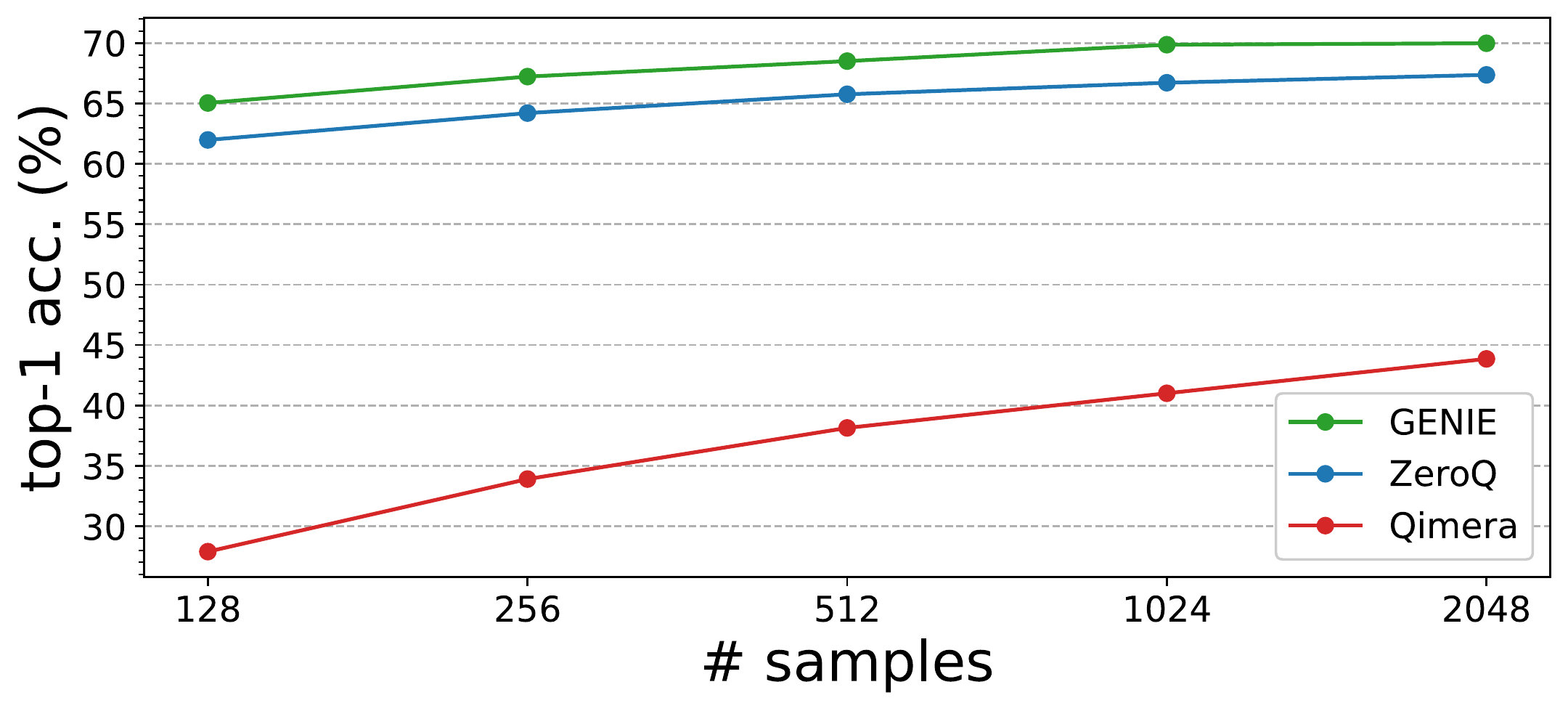}
    \caption{ResNet-50}
    \label{fig:num_b}
  \end{subfigure}
  \hfill
  \begin{subfigure}{0.95\linewidth}
    \centering
    \includegraphics[width=1.0\linewidth]{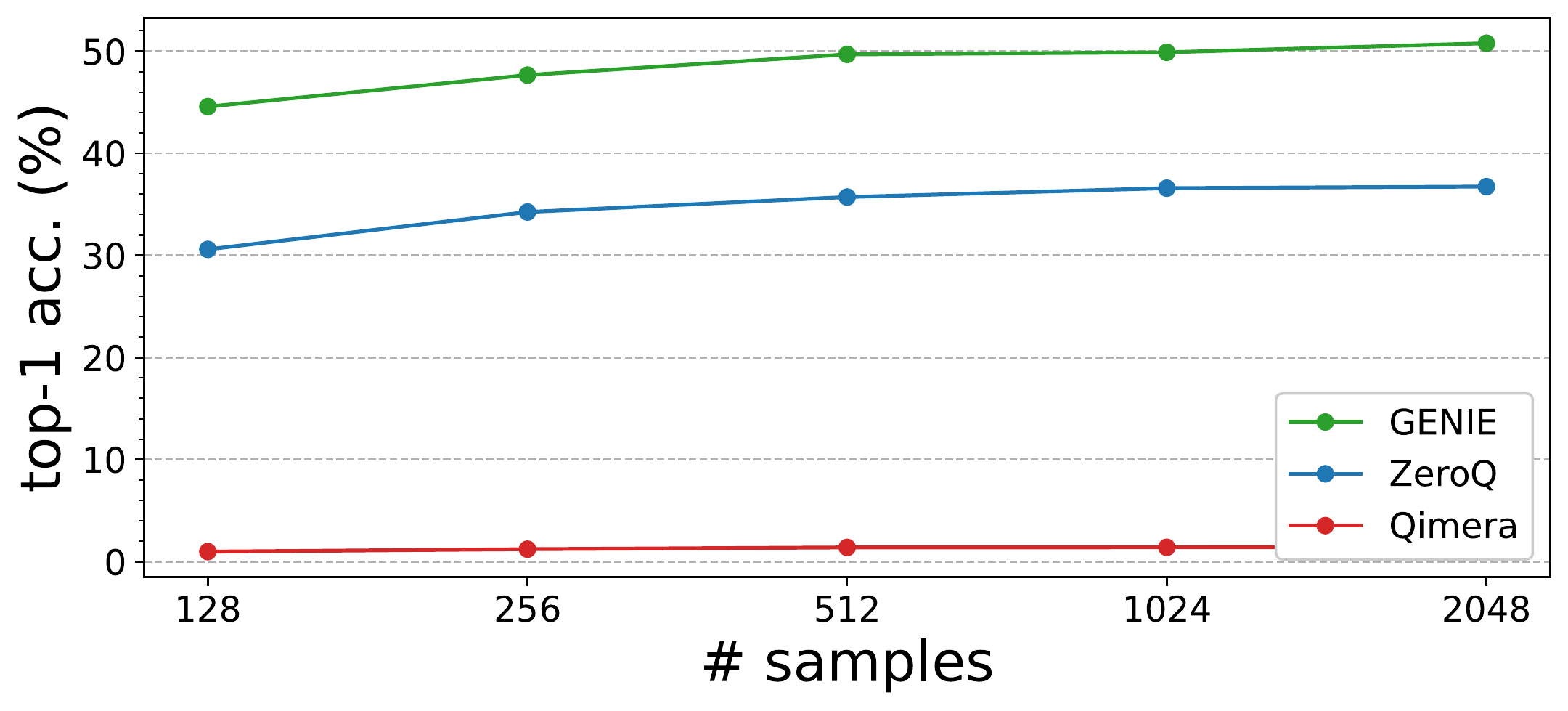}
    \caption{MobileNetV2}
    \label{fig:num_c}
  \end{subfigure}
  \caption{The influence of the number of samples on model accuracy (W2A4)}
  \label{fig:num_sample_a}
\end{figure}

\section{Generator}

The generator used in \genie-D is modified based on the generator of GDFQ~\cite{xu2020generative}, and it accepts latent vectors of size 256 as inputs. 
To reduce dependency on the generator, we use only one upsampling block, which performs the following sequence of operations: "Upsampling-Conv2D-BatchNorm-LeakyReLU". In contrast, GDFQ uses two upsampling blocks with latent vectors of size 100 as inputs (Figure~\ref{fig:gen}).
Intuitively, increasing the size of the latent vectors could produce diverse data while a deeper generator could help in learning more \textit{common knowledge} of the input domain. 
However, the performance of the quantized models does not highly depend upon the depth of the generator and the size of the latent vectors in our experiments.

\section{Informativeness of Synthetic Data}

To measure the informativeness of the synthetic data, we conduct experiments on the influence of the number of samples on model accuracy, where we identify how much information the distilled data provide when quantizing networks.
As shown in Table~\ref{tab:num_sample} and Figure~\ref{fig:num_sample_a} (where we use \textsc{QDrop} as the quantizer), the distilled data by \genie-D is more contributing to the enhancement of quantized networks.
Especially, the quantization results with 128 images from \genie-D outperform those with 1K images from Qimera throughout all models. 
In other words, a small quantity of images produced by \genie-D provides more helpful information with quantized networks, compared with other methods.
Thus, we can consider that the images by \genie-D are more informative to model quantization than those by other methods.

\begin{figure}[t]
  \centering
  \begin{subfigure}{1.0\linewidth}
    \centering
    \includegraphics[width=\linewidth]{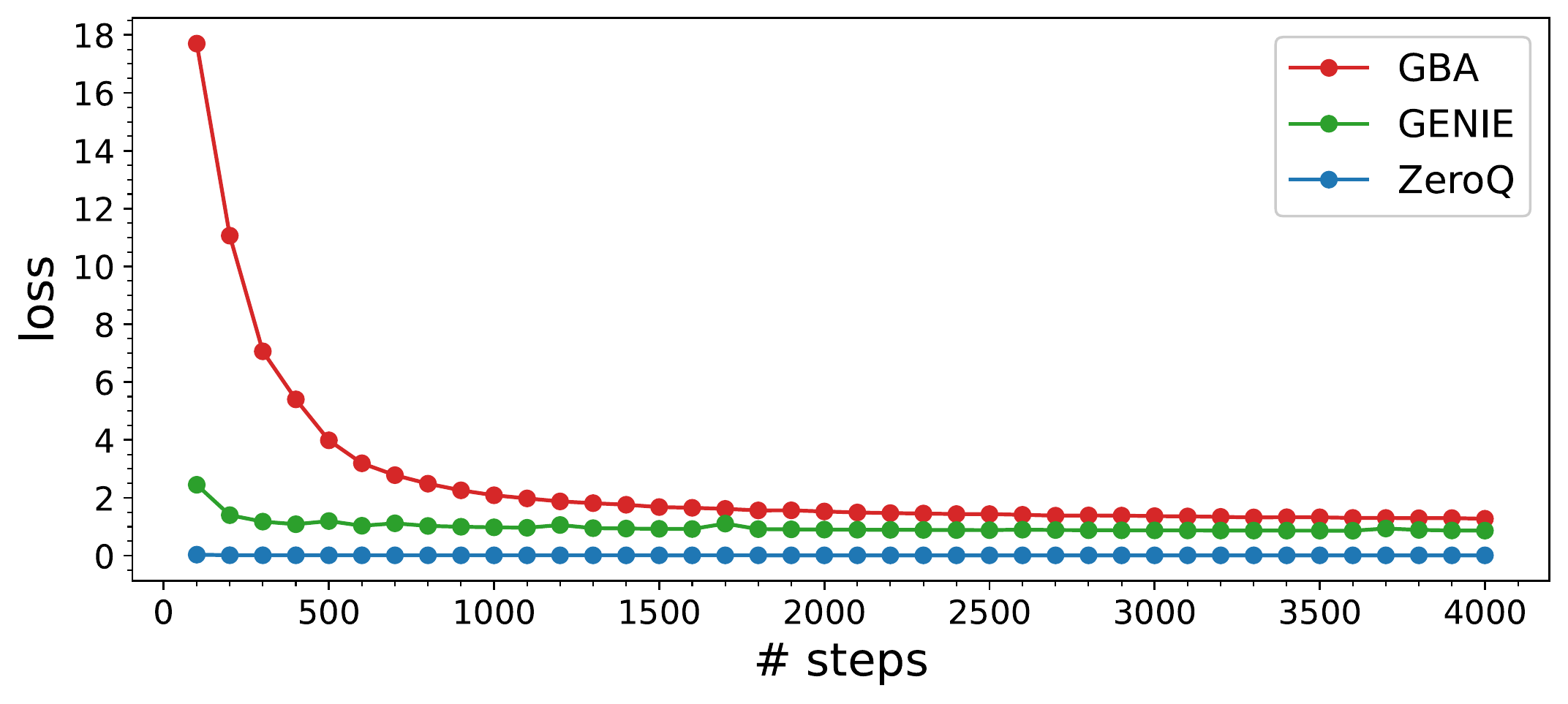}
  \end{subfigure}
  \hfill
  \begin{subfigure}{1.0\linewidth}
    \centering
    \includegraphics[width=1.0\linewidth]{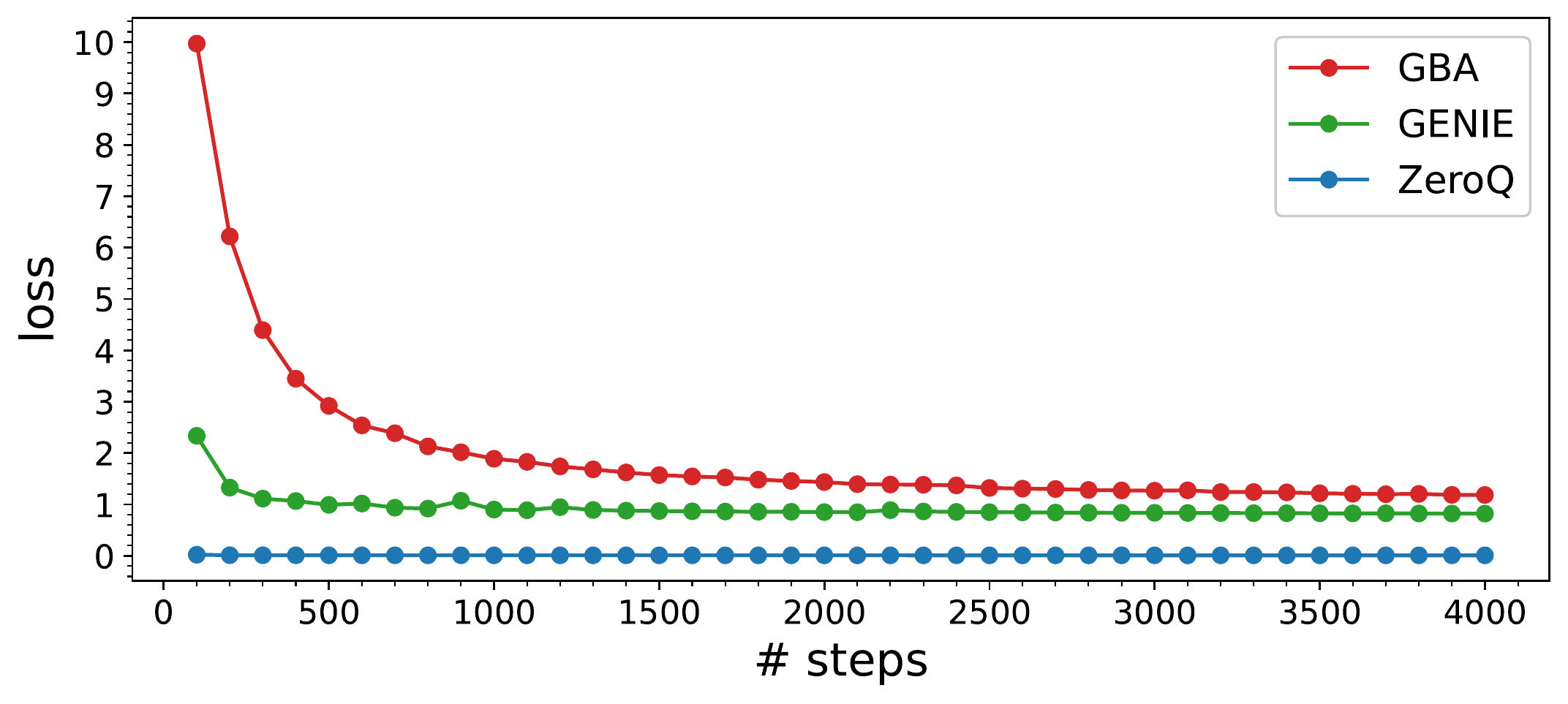}
  \end{subfigure}
  \hfill
  \begin{subfigure}{1.0\linewidth}
    \centering
    \includegraphics[width=1.0\linewidth]{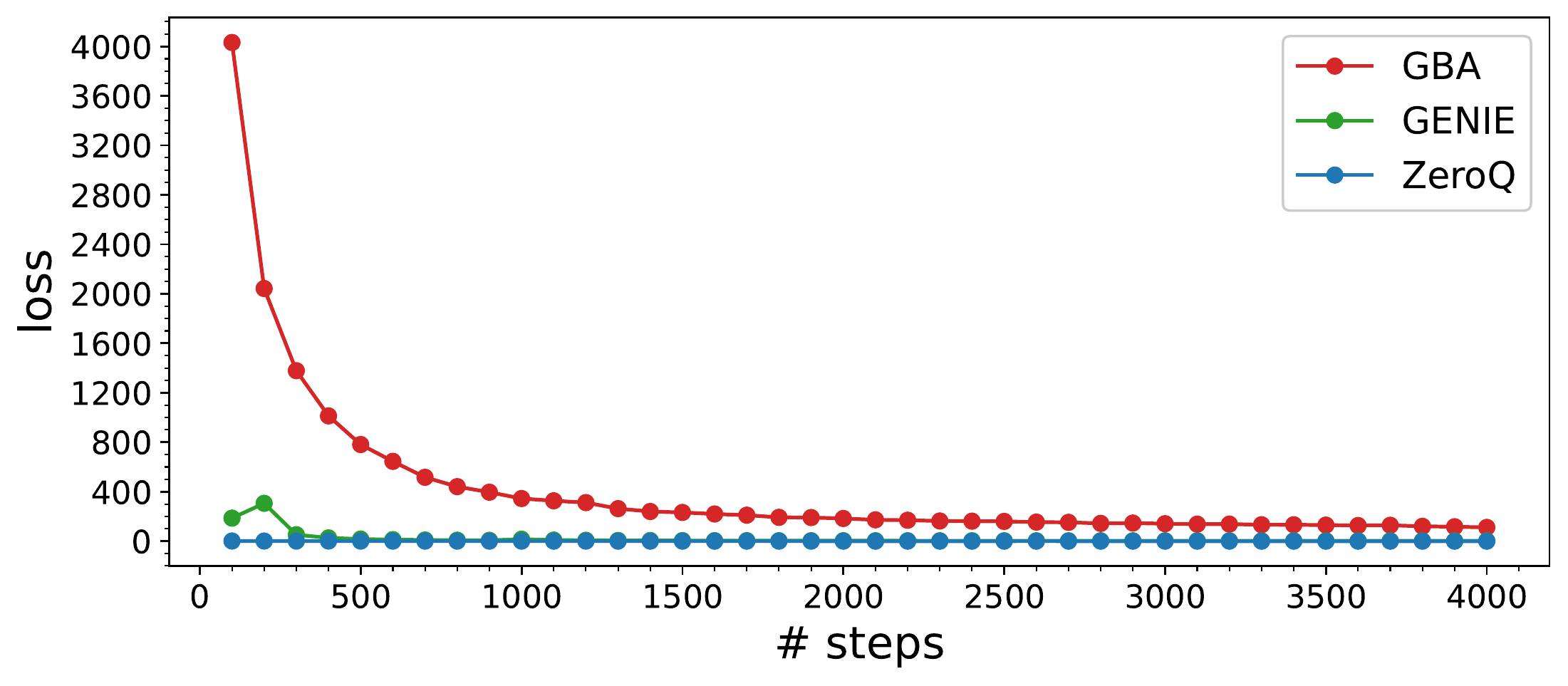}
  \end{subfigure}
  \caption{The trace of the BNS loss for the three approaches: ZeroQ, GBA, and \genie}
  \label{fig:gen_cvg}
\end{figure}

\section{Comparison on Convergence}

Figure~\ref{fig:gen_cvg} compares the trace of the BNS loss (Eq.~(\ref{eq:BNS})) for the three approaches: ZeroQ distills knowledge to the images directly. GBA uses the generator using the Gaussian noise as the input to synthesize samples. \genie-D distills knowledge into the latent vectors while optimizing the generator. 
Unlike GBA, by training latent vectors, the loss of \genie-D converges to a lower loss than that of GBA; however, it is not lower than that of ZeroQ in spite of its better quantization performance. This implies that learning \textit{common knowledge} or image prior by the generator is as important as achieving a low loss.

\begin{table}[tb]
\centering
\footnotesize
\begin{tabular}{ccccc}
\hline
\multicolumn{1}{c}{} & \multicolumn{1}{c}{Methods} & \multicolumn{1}{c}{\begin{tabular}[c]{@{}c@{}}\#Bits\\ (W/B)\end{tabular}} & \multicolumn{1}{c}{\begin{tabular}[c]{@{}c@{}} \#Synthetic\\ dataset\end{tabular}} & \multicolumn{1}{c}{\begin{tabular}[c]{@{}c@{}}Top-1\\ Accuracy(\%)\end{tabular}} \\ \hline
- &   Full Prec.  & 32/32 & - & 71.47 \\ \hline                     
\multirow{8}{*}{QAT}    &  GDFQ+AIT &\multirow{9}{*}{4/4}   &  & 65.51 \\
& ARC+AIT &   & $1.28M$ & 65.73 \\
& Qimera+AIT  &   &  & 66.83 \\  \cdashline{2-2} \cdashline{4-5}
&   \multirow{5}{*}{\textsc{Genie}-D+AIT} &  & $1K$ & 63.98 \\
&   &   & $5K$   & 65.88 \\
&   &   & $10K$   & 66.55 \\
&   &   & $20K$   & 66.67 \\
&   &   & $100K$  & 66.91 \\ \cline{1-2} \cline{4-5}
{PTQ} &   \textsc{Genie} [ours]    &   & $1K$  & \textbf{68.51} \\  \hline
\end{tabular}
\caption{\label{tab:qatptq} Comparison between PTQ and QAT on ResNet-18 \\ }
\end{table}

\section {PTQ vs. QAT}

To evaluate the performance of the data distilled by \genie-D with QAT, we adopt AIT as the quantizer and vary the number of samples to identify how much data to need for QAT. 
As shown in Table~\ref{tab:qatptq}, the size of the synthetic data does not significantly affect the performance of quantized networks. As well, the results show poor performance rather than that of PTQ with only $1K$ images.
Considering both the performance and time for image generation and training (including time for hyperparameter searching), PTQ is more efficient and suitable for ZSQ. 
Existing works train only the generator, and thus can generate data infinitely. Because they use $80K$ steps with 16 batches for QAT, they generate a total of $1.28M$ during the training.

\end{document}